\documentclass[twocolumn]{svjour3}
\smartqed  
\usepackage{booktabs}
\usepackage{graphicx}
\usepackage{amsmath}
\usepackage{amssymb}
\usepackage{tabularx}
\usepackage{multirow}
\usepackage{algorithm}
\usepackage{algorithmic}
\usepackage{stackengine}
\usepackage{xcolor}
\usepackage{adjustbox}
\usepackage{colortbl}
\usepackage{ragged2e}
\usepackage{rotating,soul}
\usepackage[caption=false,font=footnotesize]{subfig}
\usepackage[left=1.8cm,right=1.8cm,top=2.0cm,bottom=2.7cm]{geometry}
\usepackage{tipa}
\usepackage{ulem}
\usepackage{color}
\usepackage{float}
\usepackage{bm}
\usepackage{bbding}
\usepackage{verbatim}
\usepackage{makecell}
\usepackage{overpic}
\usepackage{setspace}
\usepackage[misc]{ifsym}
\usepackage{enumerate}
\usepackage{pifont}
\usepackage{bbm}
\usepackage{array}
\usepackage{fixltx2e}
 \usepackage{enumitem}
\usepackage[numbers,sort]{natbib}
\bibpunct[,]{(}{)}{;}{a}{}{,}

\usepackage{pifont}%
\newcommand{\cmark}{\ding{51}}%
\newcommand{\xmark}{\ding{55}}%

\definecolor{mygray}{gray}{.93}
\definecolor{mygray1}{gray}{.99}
\definecolor{mygray2}{gray}{.93}
\definecolor{darkpink}{rgb}{0.91, 0.33, 0.5}
\usepackage{bbding}
\usepackage{stfloats}

\def\ie{\textit{i.e.}}
\def\eg{\textit{e.g.}}
\def\etc{\textit{etc}}

\definecolor{linkcolor}{RGB}{255,0,0}
\definecolor{urlcolor}{RGB}{255,105,180}
\definecolor{citecolor}{RGB}{0, 80, 200}

\definecolor{darkpastelgreen}{rgb}{0.01, 0.75, 0.24}
\definecolor{darkpastelred}{RGB}{230, 50, 0}

\usepackage{hyperref}
\hypersetup{colorlinks=true,linkcolor=linkcolor,urlcolor=urlcolor,citecolor=citecolor}

\journalname{International Journal of Computer Vision}

\begin{document}
%
\title{Grounded Affordance from Exocentric View}
%
%
%
%

\titlerunning{Grounded Affordance from Exocentric View}        
	\author{
		Hongchen Luo$^{\textbf{1}\textbf{*}}$\and
		Wei Zhai$^{\textbf{1}\textbf{*}}$\and 
		Jing Zhang$^\textbf{2}$\and 
		Yang Cao$^\textbf{1,3 \Letter}$ \and \\
		Dacheng Tao$^{\textbf{2}}$
	}
	
	\authorrunning{Luo et al.} %

	\institute{
          Hongchen Luo (lhc12@mail.ustc.edu.cn) \\
	   Wei zhai (wzhai056@ustc.edu.cn) \\
       Jing Zhang (jing.zhang1@sydney.edu.au) \\
       \Letter~Yang Cao (forrest@ustc.edu.cn)\\
       Dacheng Tao (dacheng.tao@gmail.com) \\
		$^{\textbf{1}}$University of Science and Technology of China, Hefei, China \\
		$^{\textbf{2}}$The University of Sydney, Sydney, Australia \\
		$^{\textbf{3}}$Institute of Artificial Intelligence, Hefei Comprehensive National Science Center \\
		$^{\textbf{*}}$Hongchen Luo and Wei Zhai contributed equally. \\
	}

\date{Received: date / Accepted: date}

\maketitle

\begin{abstract}
Affordance grounding aims to locate objects' ``action possibilities'' regions, an essential step toward embodied intelligence. Due to the diversity of interactive affordance, \ie, the uniqueness of different individual habits leads to diverse interactions, which makes it difficult to establish an explicit link between object parts and affordance labels. Human has the ability that transforms various exocentric interactions into invariant egocentric affordance to counter the impact of interactive diversity. To empower an agent with such ability, this paper proposes a task of affordance grounding from the exocentric view, \ie, given exocentric human-object interaction and egocentric object images, learning the affordance knowledge of the object and transferring it to the egocentric image using only the affordance label as supervision. However, there is some ``interaction bias'' between personas, mainly regarding different regions and views. To this end, we devise a cross-view affordance knowledge transfer framework that extracts affordance-specific features from exocentric interactions and transfers them to the egocentric view to solve the above problems. Furthermore, the perception of affordance regions is enhanced by preserving affordance co-relations. In addition, an affordance grounding dataset named AGD20K is constructed by collecting and labeling over 20K images from $36$ affordance categories. Experimental results demonstrate that our method outperforms the representative models regarding objective metrics and visual quality. The code is available via: \href{https://github.com/lhc1224/Cross-view-affordance-grounding}{github.com/lhc1224/Cross-View-AG}.

\end{abstract}

\begin{figure}[t]
	\centering
		\begin{overpic}[width=0.94\linewidth]{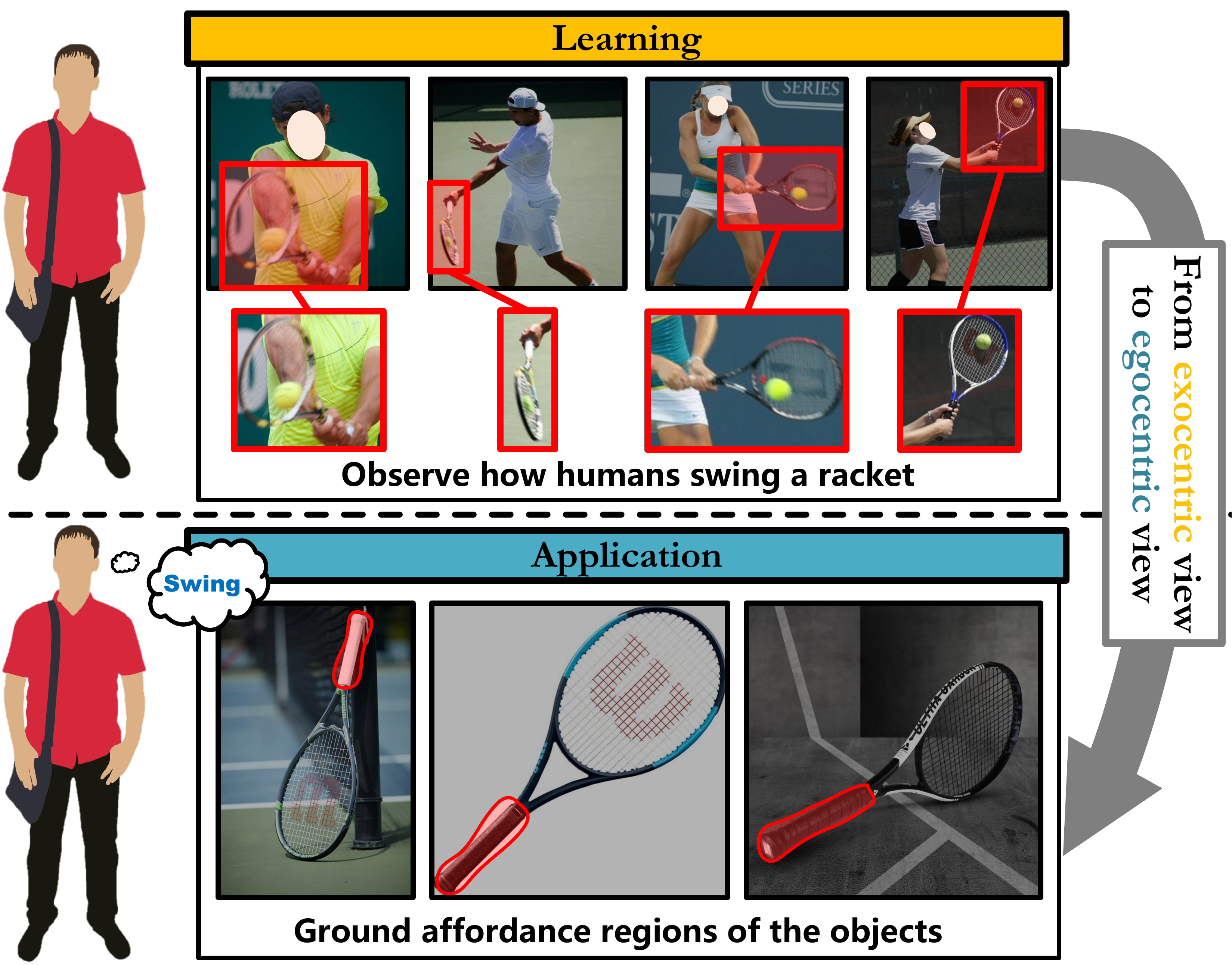}
	\end{overpic}
	\caption{\textbf{Observation.} By observing the exocentric diverse interactions, the human learns affordance knowledge determined by the object's intrinsic properties and transfers it to the egocentric view.}
	\label{figure1}
\end{figure}

\section{Introduction}
\label{sec:introduction}
Affordance grounding aims to locate an object's region of ``action possibilities''. For an intelligent agent, it is necessary to know not only what the object is but also to understand how it can be used \citep{gibson1977theory}. Perceiving and reasoning about possible interactions in local regions of objects is the key to the shift from passive perception systems to embodied intelligence systems that actively interact with and perceive their environment \citep{bohg2017interactive,nagarajan2020learning}. It has a wide range of applications for robot grasping, scene understanding, and action prediction \citep{mandikal2021learning,hassanin2021visual,grabner2011makes,koppula2013learning,zhang2020empowering,luo2021learning,yang2021collaborative}.
\par Affordance is a dynamic property closely related to the interaction between humans and environments \citep{hassanin2021visual}. As shown in Fig. \ref{figure1}, the uniqueness of different individual habits leads to diverse interactions, making it difficult to understand how to interact with objects and establish links between object parts and affordance labels \citep{luo2021one}. In contrast, humans can easily perceive the object's affordance region by observing diverse exocentric human-object interactions and giving a unique egocentric definition. Although different persons hold the racket in different positions due to their habits, the observer can distinguish swingable regions determined by the intrinsic properties (such as the long handle structure) of the racket from a collection of interacting images and transfer the knowledge to the egocentric view, thereby creating a bridge between the object part and the affordance category.

\begin{figure}[t]
	\centering
		\begin{overpic}[width=0.94\linewidth]{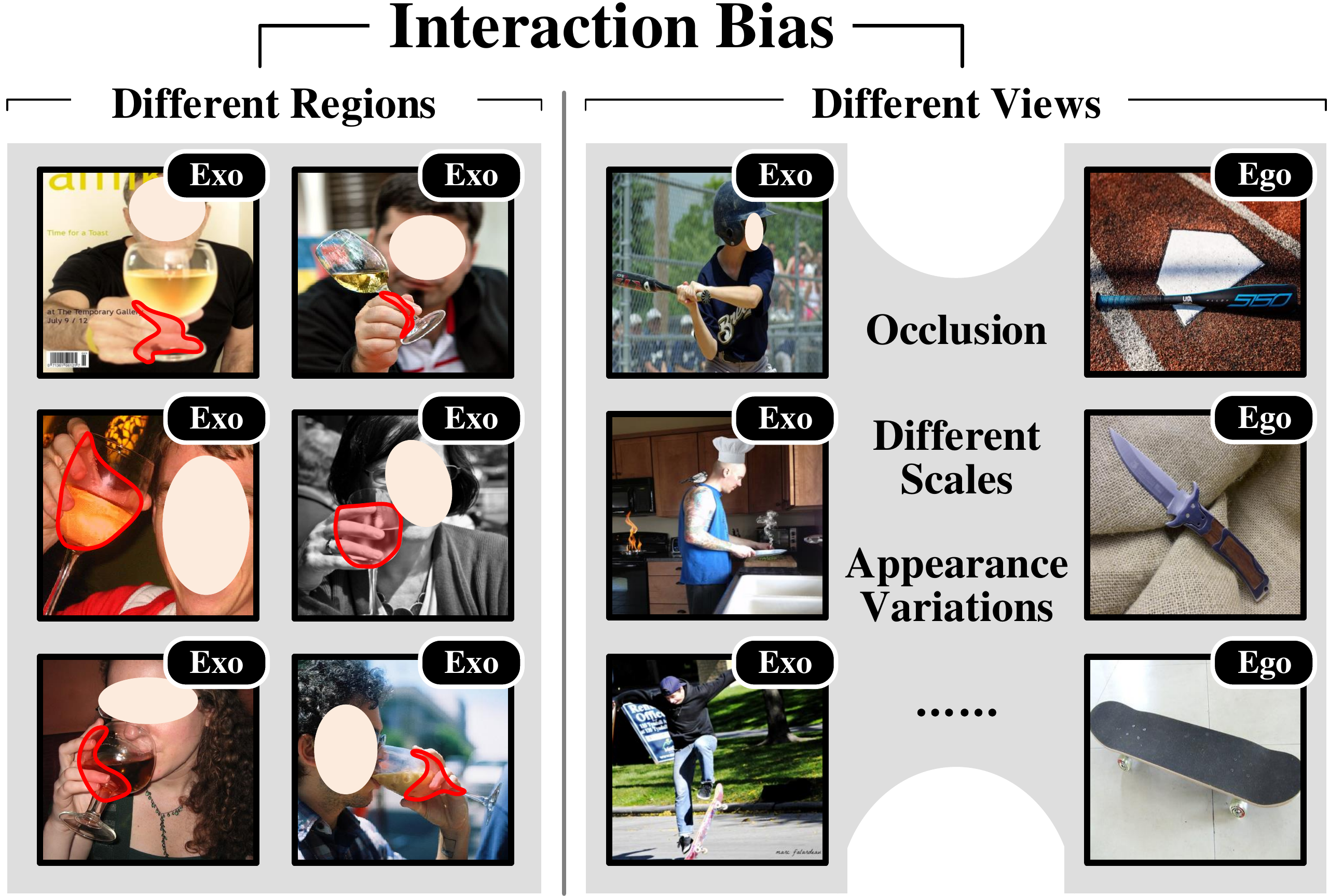}
      \put(19,-5){\textbf{(a)}}
      \put(69,-5){\textbf{(b)}}
	\end{overpic}
        \vspace{7pt}
	\caption{\textbf{The main challenge is the ``interaction bias'' between the personas.} \textbf{(a) ``interaction bias'' between regions}, \ie different habits leading to regional variations in interaction. \textbf{(b) ``interaction bias'' between views}, due to occlusion, different scales, and apparent variations, the affordance regions of the two views cannot align directly.
	}
	\label{challenge}
\end{figure}

\par To empower an agent with this ability to perceive the invariant egocentric affordance from various exocentric interactions, this paper first proposes a task of affordance grounding from the exocentric view, \ie, given exocentric human-object interactions and egocentric images, learning affordance knowledge and transferring it to object images by only using affordance labels as supervision. During testing, the model predicts the affordance region for a specific object with the input of an egocentric image and a particular affordance label. 

\begin{figure*}[t]
	\centering
		\begin{overpic}[width=0.97\linewidth]{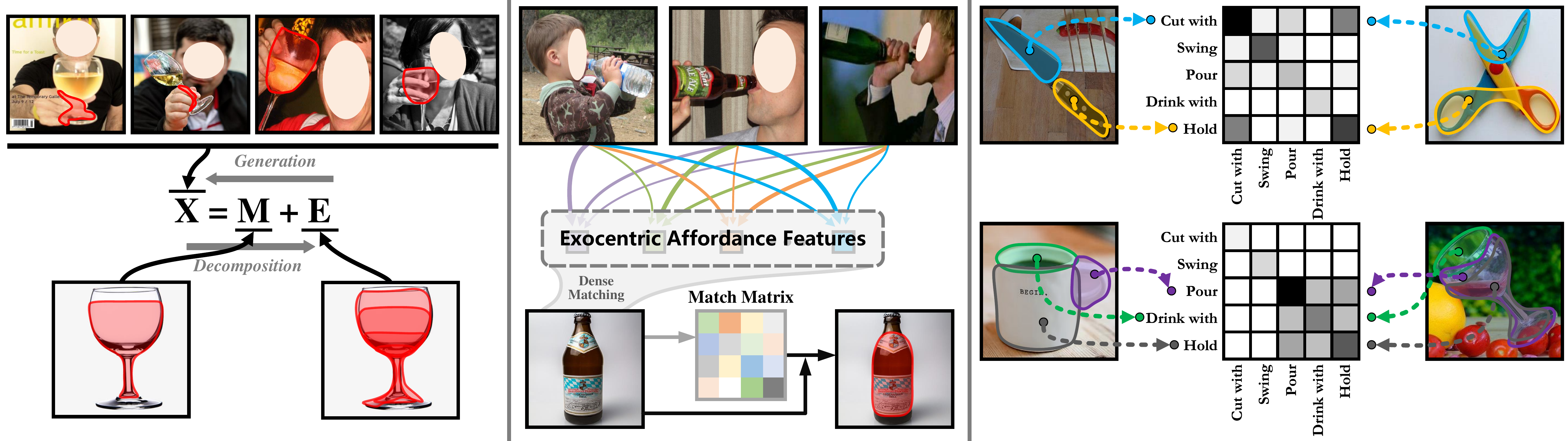}
		\put(15,-0.5){\textbf{(a)}}
		\put(46,-0.5){\textbf{(b)}}
		\put(80,-0.5){\textbf{(c)}}
	\end{overpic}
	\caption{\textbf{Motivation.} 
	 \textbf{(a)} Exocentric interactions can be decomposed into affordance-specific features $M$ and differences in individual habits $E$ (Sect. \ref{AIMmodule}). \textbf{(b)} The model obtains the cross-view match matrix by densely matching the exocentric affordance features with each egocentric location feature and thus
	  obtains the egocentric affordance feature based on the match matrix in an attentive manner 
	 (Sect. \ref{Cross-view Feature Transfer Module}). \textbf{(c)} There are co-relations between affordances which are very common due to the multiple affordance properties of objects.
	 This paper enhances the network's ability to perceive affordance regions by aligning co-relations between views (Sect. \ref{ACP}).
	}
	\label{motivation}
\end{figure*}

Bringing this power to real-world scenes would be a major leap forward, but doing so includes an issue with interpersonal ``\textbf{interaction bias}'' that has both subjective and objective aspects (as shown in Fig. \ref{challenge}). The subjective aspect is due to differences in individual habits leading to diverse interaction regions, making it challenging to locate the affordance region accurately, as shown in Fig \ref{challenge} (a). Despite such diversity, examining multiple human-object interaction images enables an exploration of objects' generic affordance regions. Thus, it is viable to disintegrate different interactions into the affordance regions dictated by objects' intrinsic features and human habit disparities, as shown in Fig. \ref{motivation} (a). This paper employs non-negative matrix factorization (NMF) methodology \citep{Lee2000AlgorithmsFN}, to reduce the variability of human habits and obtain affordance features. The objective aspect refers to the occlusion, appearance, and scale differences, making it difficult to directly align the features in the affordance region between views (as shown in Fig. \ref{challenge} (b)). This paper considers obtaining the cross-view matching matrix by densely measuring the similarity between the affordance-related exocentric feature and the egocentric feature. Then, adaptively adjust it to acquire the affordance feature representation in the egocentric view (as shown in Fig. \ref{motivation} (b)). Furthermore, there is a co-relation between affordance categories, which is independent of the semantic classes of objects (as shown in Fig. \ref{motivation} (c)). This paper enhances the network's ability to perceive affordance regions by preserving the co-relation between affordance categories.

In this paper, we propose a \textbf{cross-view affordance knowledge transfer framework}. First, an \textbf{A}ffordance \textbf{I}nvariance \textbf{M}ining (\textbf{AIM}) module is introduced to extract affordance regions from diverse exocentric interaction regions by employing non-negative matrix factorization. Then, a \textbf{C}ross-view \textbf{F}eature \textbf{T}ransfer (\textbf{CFT}) module is introduced to transfer them to the egocentric view by densely matching. Finally, an \textbf{A}ffordance \textbf{C}o-relation \textbf{P}reserving (\textbf{ACP}) strategy enhances the network's ability to perceive affordance regions by aligning the co-relation of affordance categories from both views. Specifically, the AIM module decomposes the human-object interaction into affordance-related features $M$ and personal habit differences $E$. The non-negative matrix factorization technique minimizes $E$ to obtain exocentric affordance features. In alternating iterations, the dictionary bases of the AIM module store the affordance features. The CFT module computes the cross-view matching matrix by evaluating the similarity between the dictionary bases and the features associated with each egocentric pixel. The bases and matrix are updated to accommodate differences between exocentric and egocentric views by adapting them based on egocentric features, which enables the transfer of exocentric affordance knowledge to the egocentric branch. Finally, the ACP strategy uses cross-entropy loss to align the co-relation matrices of the exocentric and egocentric branches to enhance the network's ability to perceive and locate affordance regions.

Despite the advances in affordance learning, the existing datasets \citep{Sawatzky_2017_CVPR,nguyen2017object,myers2015affordance,fang2018demo2vec} still bear limitations in terms of affordance/object category, image quality, and scene complexity. To carry out a comprehensive study, this paper presents an affordance grounding dataset named \textbf{AGD20K}, consisting of $20,061$ exocentric images and $6,060$ egocentric images from $36$ affordance categories. A comparative analysis of the AGD20K dataset uses eight prominent models across four related domains. The outcomes illustrate the superior efficacy of our approach in apprehending the intrinsic characteristics of objects and reducing the variability of affordance interaction. In summary, our primary contributions are:
\begin{itemize}
    \item [\textbf{1)}]
    We present an affordance grounding task from the exocentric view and establish a large-scale AGD20K benchmark to facilitate the research for empowering the agent to capture affordance features from exocentric interactions.
    \item [\textbf{2)}]
    We propose a novel cross-view affordance knowledge transfer framework in which the affordance knowledge is acquired from exocentric human-object interactions and transferred to egocentric views while preserving the correlation between affordances, thus achieving better perception and localization of interactive affordance.
    \item [\textbf{3)}]
    Experiments on the AGD20K dataset demonstrate that our method outperforms representative methods in several related fields and can serve as a strong baseline for future research.
\end{itemize}
This paper builds upon our conference version \citep{luo2022learning}, which has been extended in three distinct aspects. \textbf{Firstly}, we provide a deeper insight into the problem of interpersonal ``interaction bias'' in the task caused by different regions and views. \textbf{Secondly}, we introduce a cross-view feature transfer module to effectively align affordance knowledge under the egocentric view by dense comparison. \textbf{Thirdly}, we extend the dataset from multiple aspects and conduct more experiments regarding multiple attributes to comprehensively analyze the model's performance.

\par The remainder of this paper is organized as follows: Sect. \ref{sec:related work} provides a brief review of existing related studies. Sect. \ref{sec:method} describes the pipeline of the proposed model and its details. In Sect. \ref{sec:dataset}, we introduce the collection, annotation process, and statistical analysis of the AGD20K dataset. Sect. \ref{sec:experiments} describes the experimental setting and provides comprehensive results and analysis. In Sect. \ref{sec:Conclusion and Discussion}, we present the conclusions, limitations, and potential applications of this work.

\section{Related Work}
\label{sec:related work}
\subsection{Visual Affordance Learning}
Visual affordance research regards affordance perception as a computer vision issue that relies on images or videos. It employs machine learning or deep learning techniques to detect, segment, or ground ``action possibilities'' regions on objects \citep{hassanin2018visual}. Table \ref{table:affSummary} lists recent works in affordance classification, detection, segmentation, and reasoning. Numerous previous studies \citep{nguyen2017object,do2018affordancenet,chuang2018learning,fang2018demo2vec,zhao2020object} primarily rely on supervised methods to create connections between local regions of objects and their corresponding affordances. \cite{Sawatzky_2017_CVPR,sawatzky2017adaptive} achieve weakly supervised affordance detection using only a few key points.  \cite{deng20213d} expand affordance detection to 3D scenes. \cite{luo2021one,zhaione} explore human purpose-driven object affordance detection in unseen scenarios.  \cite{mi2019object,mi2020intention} and \cite{lu2022phrase} investigate affordance detection/segmentation in multimodal scenes.
\cite{nagarajan2019grounded} exploit only affordance labels to ground the interactions from the videos. In contrast to \citep{nagarajan2019grounded}, our goal is to empower the agent to learn affordance knowledge from exocentric human-object interactions. To this end, we propose an explicit cross-view affordance knowledge transfer framework that extracts affordance knowledge determined by the intrinsic properties of objects from multiple exocentric interactions and transfers it into egocentric images.
\begin{table*}[t!]
  \centering
  \scriptsize
  \renewcommand{\arraystretch}{1.}
  \renewcommand{\tabcolsep}{1.5pt}
  \caption{\textbf{Comparison of affordance-related works.} \textbf{Interaction:} the manner in which the drive model discovers affordance. \textbf{View:} the viewpoint of the input data. \textbf{CV:} cross-view. \textbf{SL:} supervised learning. \textbf{US:} whether valid on new unseen objects. \textbf{Rep:} representation of affordance.}\label{table:affSummary}
  \vspace{-1pt}
  \begin{tabular}{c||ccccccccc}
\hline
\Xhline{2.\arrayrulewidth}
 \textbf{Paper}  & \textbf{Interaction} & \textbf{View} & \textbf{CV} & \textbf{SL} & \textbf{US} & \textbf{Rep.} &  \textbf{Dataset} & \textbf{Format} & \textbf{Task} \\
\hline
\Xhline{2.\arrayrulewidth}
\cite{stark2008functional}  & None & Exo & {\color{darkpastelred} \xmark} & Fully & {\color{darkpastelred} \xmark} & Obj & ETHZ Shape  & 2D & Detection \\ 
\cite{kjellstrom2011visual} & Vision & Exo & {\color{darkpastelred} \xmark} & Fully & {\color{darkpastelred} \xmark} & - &  NORB  & 2D & Classification  \\
\cite{koppula2014physically}  &  Vision & Exo & {\color{darkpastelred} \xmark} & Fully & {\color{darkpastelred} \xmark} & Obj \& Tra  & CAD-120  & RGBD & Grounding \\
 \cite{fouhey2015defense} &  Vision & Exo & {\color{darkpastelred} \xmark} & Fully  & {\color{darkpastelred} \xmark} & Scene & NYUv2, UIUC & RGBD  & Segmentation \\
  \cite{myers2015affordance} & None & Exo & {\color{darkpastelred} \xmark} & Fully & {\color{darkpastelred} \xmark} & Part &  UMD, IIT-AFF   & RGBD & Segmentation \\
  \cite{nguyen2016detecting} & None & Exo & {\color{darkpastelred} \xmark} & Fully & {\color{darkpastelred} \xmark} & Part &  UMD, IIT-AFF  & RGBD & Segmentation \\
 \cite{do2018affordancenet} & None & Exo & {\color{darkpastelred} \xmark} & Fully & {\color{darkpastelred} \xmark} & Part &  UMD, IIT-AFF  & RGBD & Segmentation \\
 \cite{zhao2020object} & None & Exo & {\color{darkpastelred} \xmark} & Fully & {\color{darkpastelred} \xmark} & Part &  UMD, IIT-AFF  & RGBD & Segmentation \\
 \cite{lakani2017can}  & None & Exo & {\color{darkpastelred} \xmark} & Fully & {\color{darkpastelred} \xmark} & Part &  UMD, IIT-AFF  & RGBD & Segmentation \\
  \cite{srikantha2016weakly} &  None & Exo & {\color{darkpastelred} \xmark} & Weakly & {\color{darkpastelred} \xmark} & Part &  \cite{srikantha2016weakly} & RGBD & Segmentation \\
  \cite{Sawatzky_2017_CVPR} &  None & Exo & {\color{darkpastelred} \xmark} & Weakly & {\color{darkpastelred} \xmark} & Part &  \cite{srikantha2016weakly} & RGBD & Segmentation \\
  \cite{chuang2018learning}  & Multimodal & Ego & {\color{darkpastelred} \xmark} & Fully & {\color{darkpastelred} \xmark} & Scene & ADE-AFF & RGB & Segmentation \\
  \cite{fang2018demo2vec} &  Vision & Exo & {\color{darkpastelred} \xmark} & Fully & {\color{darkpastelred} \xmark} & Part & OPRA  & RGB & Grounding \\
  \cite{nagarajan2019grounded} &  Vision  & Exo/Ego & {\color{darkpastelred} \xmark} & Weakly & {\color{darkpastelgreen} \cmark} & Part & OPRA, EPIC  & RGB & Grounding \\
  \cite{luo2021learning} &  Vision  & Exo/Ego & {\color{darkpastelred} \xmark} & Weakly & {\color{darkpastelgreen} \cmark} & Part & OPRA, EPIC  & RGB & Grounding \\
  \cite{deng20213d} &  None & Exo & {\color{darkpastelred} \xmark} & Fully  & {\color{darkpastelred} \xmark} & Point cloud \& Part &  3D AffordanceNet & 3D  & Segmentation  \\
  \cite{luo2021one} &  Vision & Exo & {\color{darkpastelred} \xmark} & Fully & {\color{darkpastelgreen} \cmark} & Obj &  PAD  & RGB & Segmentation \\
  \cite{mi2020intention} &  Language & Exo & {\color{darkpastelred} \xmark} & Fully & {\color{darkpastelred} \xmark} & Obj  & \cite{mi2019object} & RGB & Detection \\
  \cite{mi2019object} &  Language & Exo & {\color{darkpastelred} \xmark} & Fully & {\color{darkpastelred} \xmark} & Obj  & \cite{mi2019object} & RGB & Detection \\
  \cite{lu2022phrase}  & Language & Exo & {\color{darkpastelred} \xmark} & Fully & {\color{darkpastelred} \xmark} & Obj  & PAD-L  & RGB & Segmentation \\
  \hline
\rowcolor{mygray}
  \textbf{\cite{luo2022learning} \& Ours} &  Vision & Exo/Ego & {\color{darkpastelgreen} \cmark} &  Weakly & {\color{darkpastelgreen} \cmark} & Part &  AGD20K & RGB & Grounding  \\
\hline
\Xhline{2.\arrayrulewidth}
  \end{tabular}
\end{table*}

\begin{table*}[!t]
  \centering
  \footnotesize
  \renewcommand{\arraystretch}{1.}
  \renewcommand{\tabcolsep}{4.7 pt}
   \caption{\textbf{Statistics of related datasets and the proposed AGD20K dataset.} \textbf{Part:} part-level annotation. \textbf{HQ:} high-quality annotation. \textbf{BG:} the background is fixed or from general scenarios. \textbf{Exo\&Ego:} whether to transfer from exocentric to egocentric view. $\bm{\sharp}$\textbf{Obj:} number of object classes. $\bm{\sharp}$\textbf{Aff}: number of affordance classes. $\bm{\sharp}$\textbf{Img}: number of images.}
   \vspace{-1pt}
\label{table:affSummarydataset}

\begin{tabular}{c|c||ccc|cccc|ccc}
\hline
\Xhline{2.\arrayrulewidth}
& \textbf{Dataset} & \textbf{Pub.} & \textbf{Year} & \textbf{link} & \textbf{HQ}  & \textbf{Part} & \textbf{BG} &\textbf{Exo\&Ego} &  \bm{$\sharp$}\textbf{Obj.} & \bm{$\sharp$}\textbf{Aff.} & \bm{$\sharp$}\textbf{Img.}   \\
\hline
\Xhline{2.\arrayrulewidth}
 1 & UMD \citep{myers2015affordance}  & ICRA      & 2015 & \href{http://users.umiacs.umd.edu/~amyers/part-affordance-dataset/}{Link}  & {\color{darkpastelred} \xmark}  & {\color{darkpastelgreen} \cmark} & Fixed & {\color{darkpastelred} \xmark} & 17  & 7  & 30,000 \\
 2 & \citep{Sawatzky_2017_CVPR} & CVPR & 2017 & \href{https://zenodo.org/record/495570}{Link}    & {\color{darkpastelred} \xmark} & {\color{darkpastelgreen} \cmark} & Fixed & {\color{darkpastelred} \xmark} & 17 & 7 & 3,090  \\
 3 & IIT-AFF \citep{nguyen2017object}  & IROS    & 2017 & \href{https://sites.google.com/site/ocnncrf/}{Link} & {\color{darkpastelred} \xmark} &  {\color{darkpastelgreen} \cmark} & General & {\color{darkpastelred} \xmark} & 10  & 9   & 8,835  \\
 4 & ADE-Aff \citep{chuang2018learning} & CVPR    &  2018 & \href{http://www.cs.utoronto.ca/~cychuang/learning2act/}{Link}   & {\color{darkpastelgreen} \cmark}   & {\color{darkpastelgreen} \cmark} & General & {\color{darkpastelred} \xmark}  & 150 & 7   & 10,000 \\
5 & PAD \citep{luo2021one} & IJCAI    & 2021  & \href{https://github.com/lhc1224/OSAD_Net/}{Link} & {\color{darkpastelgreen} \cmark}   & {\color{darkpastelred} \xmark} & General  & {\color{darkpastelred} \xmark}  & 72  & 31  & 4,002  \\
6 & PADv2 \citep{zhaione} & IJCV    & 2022  & \href{https://github.com/lhc1224/OSAD_Net/}{Link} & {\color{darkpastelgreen} \cmark}   & {\color{darkpastelred} \xmark} & General  & {\color{darkpastelred} \xmark}  & 103  & 39  & 30,000  \\
\hline
\rowcolor{mygray}
7 & \textbf{AGD20k (Ours)} & This Work & 2022  & \href{https://github.com/lhc1224/Cross-View-AG}{Link} & {\color{darkpastelgreen} \cmark}  & {\color{darkpastelgreen} \cmark} &   General & {\color{darkpastelgreen} \cmark}  & 50 & 36 & 26,117   \\
\hline
\Xhline{2.\arrayrulewidth}
    \end{tabular}
\end{table*}

\subsection{Visual Affordance Dataset}
Numerous datasets supporting affordance-related tasks are available as summarzied in Table \ref{table:affSummarydataset}. \cite{myers2015affordance} introduce a large-scale RGB-D dataset containing pixel-level affordance labels and corresponding ranks. \cite{nguyen2017object} construct the IIT-AFF dataset considering a more complex background of practical applications. \cite{Sawatzky_2017_CVPR} select video frames to construct a weakly supervised affordance detection dataset, using only cropped-out object regions but in inferior image quality. \cite{luo2021one} consider inference human's purpose from support images and transfers to a group of query images. Then, \cite{lu2022phrase} consider multimodal scenarios and extended the PAD dataset to a phase-based affordance detection dataset, but both failed to provide part-level affordance labels. Other affordance datasets \citep{myers2015affordance,nguyen2017object,chuang2018learning,fang2018demo2vec} suffer from the problems of small scale and low affordance/object category diversity and do not consider human actions to reason about the affordance regions. \cite{chuang2018learning} take the physical world and social norms into account and construct the ADE-Affordance dataset. However, they do not consider the requirement for an intelligent agent to observe and learn from the exocentric view and transfer it to the egocentric view.
In contrast to the above works, we explicitly consider exocentric-to-egocentric view transformations and collect a much larger scale of images, with richer affordance/object classes and part-level annotations, which are more valuable in developing affordance perception approaches towards practical real-world applications.

\begin{figure*}[t]
	\centering
		\begin{overpic}[width=0.969\linewidth]{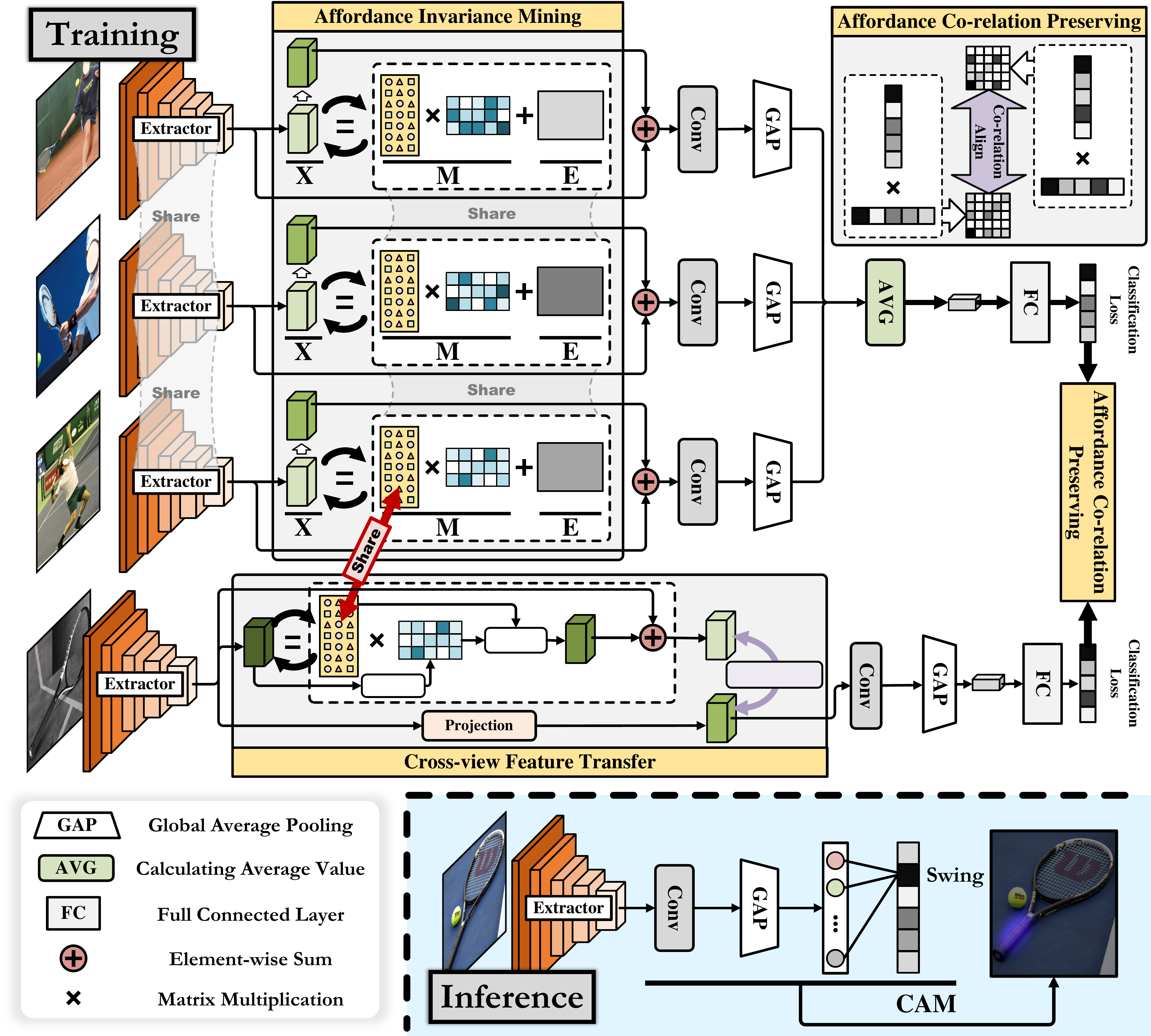}
  
		\put(27.1,73.8){\scriptsize{$\bm{1}$}}
		\put(27.1,58.6){\scriptsize{$\bm{2}$}}
		\put(26.9,43){\scriptsize{$\bm{N}$}}

		\put(39.9,43.1){\scriptsize{$\bm{N}$}}
  
		\put(50.2,42.8){\scriptsize{$\bm{N}$}}
		
		\put(39.9,58.7){\scriptsize{$\bm{2}$}}
		\put(50.2,58.7){\scriptsize{$\bm{2}$}}
		
		\put(39.9,74.0){\scriptsize{$\bm{1}$}}
		\put(50.2,74.0){\scriptsize{$\bm{1}$}}
		
		\put(33.4,84.1){$\bm{W}$}
		\put(33.43,68.3){$\bm{W}$}
		\put(33.4,53.2){$\bm{W}$}
		
		\put(40.2,51.4){$\bm{H_N}$}
		\put(40,67.1){$\bm{H_2}$}
		\put(40.2,82.3){$\bm{H_1}$}
		
		\put(27.5,83.7){$\bm{M_1}$}
   		\put(27.7,68.2){$\bm{M_2}$}
		\put(27.7,53.){$\bm{M_N}$}
		
		\put(62.6,79.6){\scriptsize{$\bm{D_1}$}}
		\put(62.6,64.5){\scriptsize{$\bm{D_2}$}}
		\put(62.4,48.9){\scriptsize{$\bm{D_N}$}}
		\put(76.5,31.6){\tiny{$\bm{D_{ego}}$}}
		
		\put(30,86){\small{$\bm{Conv}$}}
		\put(30,71.05){\small{$\bm{Conv}$}}
		\put(30,55.55){\small{$\bm{Conv}$}}
		
		
		\put(31.6,30.18){\textbf{\tiny{Eq.\ref{eq5}$ \sim $\ref{eq6}}}}
		\put(42.8,34.){\textbf{\scriptsize{Eq. \ref{eq8}}}}
		
		\put(63.55,30.85){\textbf{\scriptsize{Eq.\ref{eq11}$ \sim $\ref{eq12}}}}

         \put(48.8,30.6){\small{$\bm{M_{ego}}$}}

         \put(51.8,32){\small{$\bm{Conv}$}}

		\put(75.3,78.4){\small{\bm{$p$}}}
		\put(95.2,81){\small{\bm{$q$}}}
		
		\put(82.2,83.7){\scriptsize{\bm{$Q$}}}
		\put(87.8,71){\scriptsize{\bm{$P$}}}
		
		\put(82.,80.8){\rotatebox{270}{\textbf{\scriptsize{(Eq. \ref{LACP})}}}}
		
		\put(83.8,32.5){\bm{$d_{ego}$}}
		\put(82,65.5){\bm{$d_{exo}$}}
		\put(20.5,79.8){\bm{$Z_1$}}
		\put(20.5,64.2){\bm{$Z_2$}}
		\put(20.3,49){\bm{$Z_N$}}
		\put(15.3,34){\bm{$Z_{ego}$}}
		
		\put(89.2,57){\rotatebox{270}{\textbf{(Eq. \ref{acp1}) $ \sim $ (Eq. \ref{LACP})}}}
	\end{overpic}
	\caption{\textbf{Overview of the cross-view affordance knowledge transfer framework.} It mainly consists of an \textbf{A}ffordance \textbf{I}nvariance \textbf{M}ining (\textbf{AIM}) module (Sect. \ref{AIMmodule}), a \textbf{C}ross-view \textbf{F}eature \textbf{T}ransfer (\textbf{CFT}) module (Sect. \ref{Cross-view Feature Transfer Module}) and an \textbf{A}ffordance \textbf{C}o-relation \textbf{P}reserving (\textbf{ACP}) strategy (Sect. \ref{ACP}).}
	\label{pipeline}
\end{figure*}
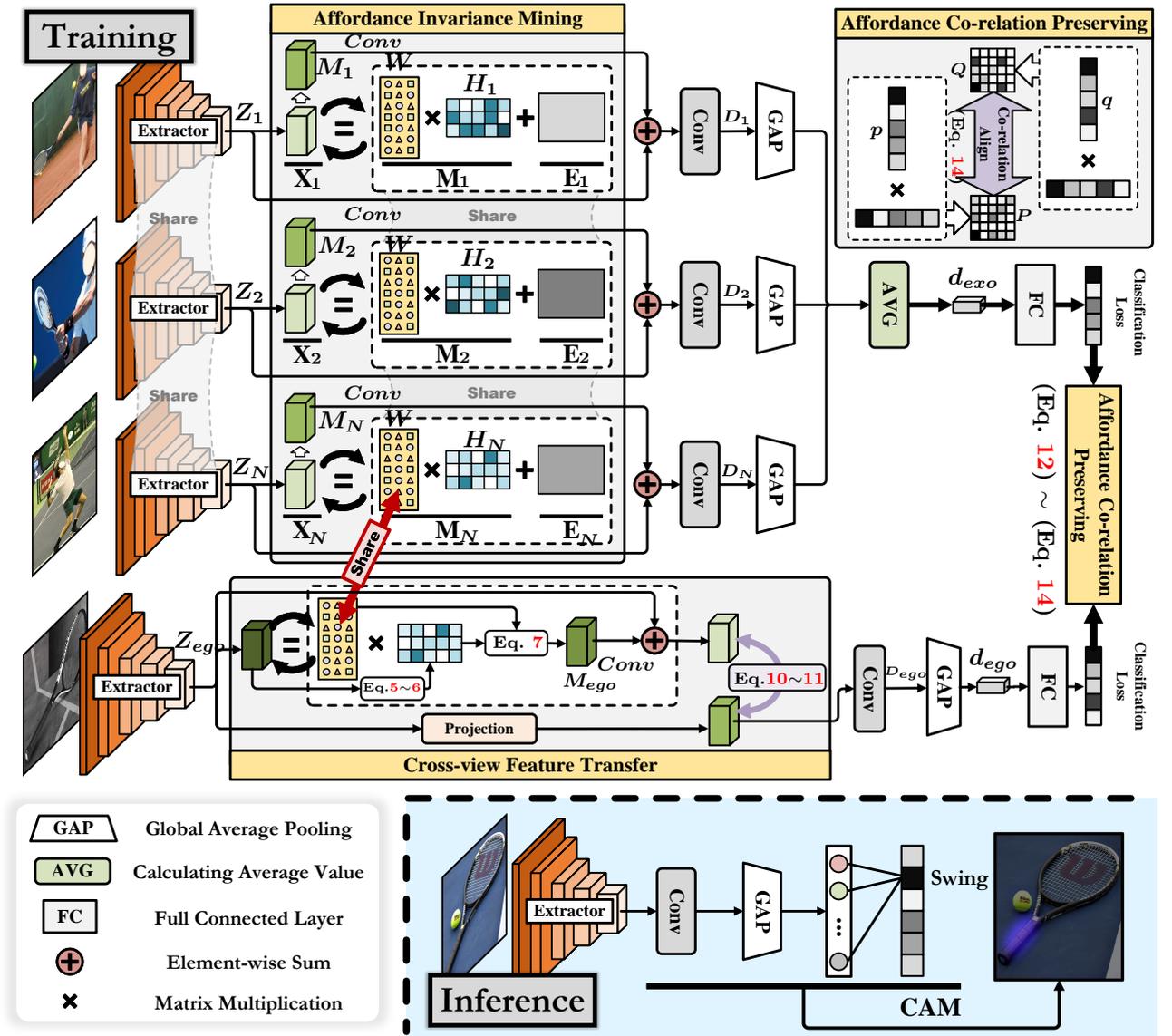

\subsection{Learning View Transformations}
The existing learning-view transformation works almost start from the mirror neurons theory \citep{rizzolatti2004mirror}, which adopts embedding learning to generate perspective invariant feature representations from paired data and then leverage it for tasks such as action recognition and video summarization under egocentric view \citep{sigurdsson2018actor,soran2014action,ho2018summarizing,regmi2019bridging,2017Identifying,lu2022learning}. \cite{sigurdsson2018actor} construct a large-scale video dataset containing first- and third-person pairs, while they use the data to learn joint embeddings in a weakly supervised setting to align the two domains, thus effectively transferring knowledge from third to first person. \cite{2017Identifying} proposes to build person-level correspondence across perspectives while introducing a novel semi-Siamese CNN architecture to address this challenge. \cite{li2021ego} extract key egocentric signals from the exocentric view dataset during pre-training and distill them to the backbone to guide feature learning in the egocentric video task. In contrast to the above works, we aim to extract affordance knowledge from the diverse exocentric human-object interactions and transfer it to the egocentric view. It is challenging due to the uncertainty caused by various interactions and objects' multiple affordance regions.

\section{Method}
\label{sec:method}
\par The goal of  the cross-view affordance grounding task is to locate the object affordance regions in egocentric images. During training, given a group of exocentric images $\mathcal{I}_{exo}=\left \{ I_1,\cdots ,I_N  \right \}$ ($N$ is the number of exocentric images) and an egocentric image $I_{ego}$, the network uses only affordance labels as supervision, to learn affordance knowledge from exocentric images and transfer it to egocentric images. During testing, only given an egocentric image $I_{ego}$ and the affordance label $C_a$, the model outputs the affordance region on the object.
Our cross-view affordance knowledge transfer framework for affordance grounding is shown in Fig. \ref{pipeline}. During training, we first use Resnet50 \citep{he2016deep} to extract the features of exocentric and egocentric images to obtain $\mathcal{Z}_{exo}=\left \{   Z_1,\cdots ,Z_N  \right \}$ and $Z_{ego}$, respectively. Then, the Affordance Invariance Mining (AIM) module is introduced to extract affordance-specific clues ($\mathcal{F}_{exo}=\left \{   F_1,\cdots ,F_N  \right \}$) from the exocentric features (see in Sect. \ref{AIMmodule}). Subsequently, the Cross-view Feature Transfer (CFT) module is proposed to transfer the affordance features extracted from the exocentric to the egocentric view (see in Sect. \ref{Cross-view Feature Transfer Module}). Afterward, the features of the two branches ($\mathcal{F}_{exo}$ and $F_{ego}$) are fed into the same convolution layer to obtain features $\mathcal{D}_{exo}$ and $D_{ego}$ respectively. We feed the $\mathcal{D}_{exo}$ through the global average pooling (GAP) layer to obtain the $d_{exo}$ and pass the $D_{ego}$ through the GAP layer to get the $d_{ego}$. Later, $d_{exo}$ and $d_{ego}$ are fed into the same fully connected layer to obtain the affordance prediction. Finally, the Affordance Co-relation Preserving (ACP) strategy is devised to enhance the network's perception of affordance by aligning the co-relation matrix of the outputs of the two views (see in Sect. \ref{ACP}). During testing, we feed the egocentric object images into the network only through the egocentric branch and then use the CAM \citep{zhou2016learning} technique to obtain the affordance regions of the object (see in Sect. \ref{Inference}). Table \ref{operation symbols} exhibits operation symbols, while Table \ref{symbols} displays symbol dimensions, domains of definition, and meanings in the methods.

\subsection{Affordance Invariance Mining Module}
\label{AIMmodule}
Human-object interaction varies across regions due to differences in human behavior, presenting challenges in obtaining complete feature representation for affordances from a single image. Nevertheless, it is possible to explore the generic features of the affordance regions from multiple human-object interaction images. These multiple interactions can be deconstructed using affordance invariant features and individual habits. The affordance invariance mining (AIM) module seeks to extract affordance invariant features from exocentric images depicting human-object interaction.

As shown in Fig. \ref{pipeline}, the exocentric interactions are decomposed into affordance-specific features $M$ and individual differences $E$. The aim is to improve the affordance feature representation $M$ by reducing individual variation in habits $E$. Inspired by low-rank matrix factorization \citep{kolda2009tensor,Lee2000AlgorithmsFN,geng2021attention}, we represent the $M$ as the multiplication of a dictionary matrix $W$ and a coefficient matrix $H$, where the dictionary bases represent the sub-features of human-object interaction and minimize $E$ by iterative optimization to obtain a reconstructed affordance representation $M$. Specifically, for the input $Z_i$, we first reduce its dimensionality with a convolution layer and a ReLU layer to ensure the non-negativity of the input and then reshape them into $X_i \in \mathbb{R}^{c \times hw}$ ($c$, $h$ and $w$ are the channels, length, and width of the feature maps respectively). We use non-negative matrix factorization (NMF) \citep{Lee2000AlgorithmsFN} to update the dictionary and the coefficient matrices. Consequently, $X_i$ is decomposed into two non-negative matrices $W$ and $H_i$. Here $W \in \mathbb{R}^{c \times r}$ is the dictionary matrix shared by all exocentric features, while $H_i \in \mathbb{R}^{r \times hw}$ is the coefficient matrix of each exocentric feature, and $r$ is the rank of the low-rank matrix $W$. To update $H_i$ and $W$ in parallel, we concatenate $\mathcal{X}_{exo}=\left \{   X_1,\cdots ,X_N  \right \}$ and $\mathcal{H}=\left \{   H_1,\cdots ,H_N  \right \}$ to obtain $X \in \mathbb{R}^{c \times Nhw}$ and $H \in \mathbb{R}^{r \times Nhw}$. The optimization process is as follows: 
\begin{equation}
\small
    \mathop{\min}\limits_{W,H}  || X-WH ||, \quad s.t. \ W_{ab}\ge 0,H_{bk} \ge 0. \label{eq1}
\end{equation}
$W$ and $H$ are updated according to the following rules:
\begin{equation}
\small
    H_{ab} \leftarrow H_{ab}\frac{\left ( W^TX \right )_{ab}}{\left ( W^TWH \right )_{ab}}, W_{ab} \leftarrow W_{ab}\frac{\left ( XH^T \right )_{ab}}{\left ( WHH^T \right )_{ab}}. \label{WH}
\end{equation}
After several iterations, we get the output $M=WH$, and reshape it to $\mathcal{M}_{exo}=\left \{   M_1,\cdots ,M_N  \right \}$. Finally, we use a convolution layer ($f$) to map it to the residual space and sum it with the $\mathcal{Z}$ to get the final output $\mathcal{F}_{exo}=\{F_1,...,F_N\}$:
\begin{equation}
\small
      F_i=\text{ReLU}\left ( Z_i+f\left ( M_i \right ) \right ), \quad i\in \left [ 1,N \right ]. \label{eq4}
\end{equation}
In each batch of training, we update the initial dictionary matrix $W^{(0)}$ such that it can accumulate the statistical prior of the common subfeature of human-object interaction, \ie,
\begin{equation}
\small
    W^{(0)} \leftarrow \alpha W^{(0)}+\left ( 1-\alpha \right )\bar{W}, \label{eq5}
\end{equation}
where $\bar{W}$ is the average over each mini-batch, and $\alpha \in [0,1]$ is the momentum, which is set to 0.9 by default.

\subsection{Cross-view Feature Transfer Module}
\label{Cross-view Feature Transfer Module}
The different views, scale, and occlusion cause difficulties in transferring affordance features from the exocentric to the egocentric view, and direct distillation may lead to lost spatial information and hard-to-perceive local detail features. Since the AIM module contains compact and comprehensive affordance region-specific detail cues after several iterations, we propose utilizing it as the initial value to densely compare the similarity of the egocentric feature with the dictionary base, ultimately achieving the cross-view matching matrix. Further, the dictionary bases are updated to adapt to the variation in viewpoint and scale in the egocentric branches by alternate iterative optimization with the egocentric features. It leads to improved optimization of the affordance region features activated in the egocentric. Finally, alignment is conducted to maintain the unique local details of the egocentric features during affordance knowledge transfer. 

The CFT modoule is shown in Fig. \ref{pipeline}. Specifically, $Z_{ego}$ is passed through the convolution layer and ReLU to ensure the non-negativity of the egocentric features $X_{ego}$. The cross-view matching matrix $H_{ego}$  is generated by performing dense comparisons between the egocentric feature pixels and the dictionary base $W$, followed by a Softmax activation for normalization:

\begin{equation}
\small
    X_{ego}=\text{ReLU}\left ( f\left ( Z_{ego} \right ) \right ), \label{eq5}
\end{equation}
\begin{equation}
\small
    H_{ego}=\text{Softmax}(X_{ego}^TW). \label{eq6}
\end{equation}

\begin{table}[!t]
\caption{Meaning of the operation}
\centering
 \small

 \renewcommand{\arraystretch}{1.}
  \renewcommand{\tabcolsep}{3 pt}
  \begin{tabular}{c|c}
\hline
\Xhline{2.\arrayrulewidth}
    \textbf{Operation} &   \textbf{Meanings}  \\
\hline
\Xhline{2.\arrayrulewidth}
  Conv $f$ & Convolution operation \\
   Max &  Take the maximum value along the channel \\
   Softmax & Softmax operation \\
   GAP &  Global average pooling \\
   ReLU & ReLU activation function \\
   Project & Project layer \\
   $||\centerdot||$ & $L_2$ loss \\
   
\hline
\Xhline{2.\arrayrulewidth}
    \end{tabular}
    \label{operation symbols}
\end{table}
Subsequently, the values of $W$ and $H_{ego}$ are alternately adjusted utilizing an iterative optimization of the NMF of Eq. \ref{WH}, enabling the adaptive tuning of the values of $W$ to the egocentric branch features. Then, the activation features $M_{ego}$ are obtained by reconstructing $W$ and $H_{ego}$. Finally, $M_{ego}$ is reshaped to $\mathbb{R}^{c \times h \times w}$ and mapped to the same dimension as $Z_{ego}$ and augmented with the $Z_{ego}$ to obtain the activation feature $\tilde{F}_{ego}$:
\begin{equation}
\small
    M_{ego}=WH_{ego}, \label{eq8}
\end{equation}
\begin{equation}
\small
    \tilde{F}_{ego}=\text{ReLU}\left ( f\left ( M_{ego} \right )+Z_{ego} \right ).
\end{equation}
To ensure that the egocentric features maintain their unique local details during the functional knowledge transfer, we perform an alignment with the egocentric branch features. Specifically, $Z_{ego}$ is inputted to the project layer  $\text{Project}\left ( \cdot \right )$ to produce the feature representation $\bar{F}_{ego}$: 
\begin{equation}
\bar{F}_{ego}=\text{Project}\left ( Z_{ego}  \right ).    
\end{equation} 
Then, the maximum response value of each pixel region is selected for alignment:
\begin{equation}
\small
    \tilde{V}_{ego}=\text{Max}\left ( \tilde{F}_{ego} \right ), \quad 
    \bar{V}_{ego}=\text{Max}\left ( \bar{F}_{ego} \right ), \label{eq11}
\end{equation}
where $\tilde{V}_{ego}$ and $\bar{V}_{ego}$ represent the channel maximum response values for each pixel of $\tilde{F}_{ego}$ and $\bar{F}_{ego}$, respectively. Finally, the alignment is calculated as the $L2$ loss between $\tilde{V}_{ego}\left ( \tilde{V}_{ego} \in \mathbb{R}^{h \times w} \right )$ and $\bar{V}_{ego}\left ( \bar{V}_{ego} \in \mathbb{R}^{h \times w} \right )$:
\begin{equation}
\small
    L_{KT}=|| \tilde{V}_{ego}-\bar{V}_{ego} ||.  \label{eq12}
\end{equation}

\begin{table}[!t]
\caption{\textbf{The dimensions, domains of definition, and meanings of the symbols used in the proposed approach. \textbf{Dim.}: Dimensions.}
}
\centering
 \scriptsize
 \renewcommand{\arraystretch}{1.}
  \renewcommand{\tabcolsep}{0.4 pt}
  \begin{tabular}{c||c|c|c}
\hline
\Xhline{2.\arrayrulewidth}
     &  \textbf{Dim.} & \textbf{Domains}  & \textbf{Meanings}  \\
\hline
\Xhline{2.\arrayrulewidth}
   $I_{i}$/$I_{ego}$ &  3$\times$224$\times$224 & $\left [ -1,1 \right ]$ & \scriptsize{Exocentric/Egocentric image} \\
   $Z_i$/$Z_{ego}$ & 2048$\times$$w$$\times$$h$ & $\left [ -\infty, +\infty \right ]$ & \scriptsize{Exocentric/Egocentric feature} \\
   $X_i$ & $c$$\times$$w$$\times$$h$ & $\left [ 0, +\infty \right ]$ & \scriptsize{$Z_i$ after dimensionality reduction} \\
   $X$ & $c$$\times$$Nhw$ & $\left [ 0, +\infty \right ]$ & \scriptsize{$X_i$ reshape after concatenating} \\
   $W$ & $c$$\times$$r$ & $\left [ 0, +\infty \right ]$ & \scriptsize{Dictionary matrix} \\
   $H$ & $r$$\times$$Nhw$ & $\left [ 0, +\infty \right ]$  & \scriptsize{Coefficient matrix} \\
   $M$ &  $c$$\times$$Nwh$ & $\left [ 0, +\infty \right ]$ & \scriptsize{Reconstructed from $W$ and $H$}  \\
   $F_i$  & $c$$\times$$Nhw$ & $\left [ 0, +\infty \right ]$ & \scriptsize{Output of the AIM module} \\
   $X_{ego}$ & $c$$\times$$wh$ & $\left [ 0, +\infty \right ]$ & \scriptsize{$Z_{ego}$ after dimensionality reduction}\\
   $H_{ego}$ & $r$$\times$$hw$ & $\left [ 0, +\infty \right ]$ & \scriptsize{Coefficient matrix for CFT module}   \\
   $M_{ego}$ & $c$$\times$$hw$ & $\left [ 0, +\infty \right ]$ & \scriptsize{Reconstructed from $W$ and $H_{ego}$} \\
   $\tilde{F}_{ego}$ & $c$$\times$$h$$\times$$w$ & $\left [ 0, +\infty \right ]$ & \scriptsize{Activate feature for egocentric} \\
   $\hat{F}_{ego}$ & $c$$\times$$h$$\times$$w$ & $\left [ 0, +\infty \right ]$ & \scriptsize{Project feature} \\
   $D_{exo}$ & 1024$\times$$h$$\times$$w$ & $\left [ -\infty, +\infty \right ]$ & \scriptsize{$F_i$ after convolution}  \\
   $D_{ego}$ & 1024$\times$$h$$\times$$w$ & $\left [ -\infty, +\infty \right ]$ & \scriptsize{$F_{ego}$ after convolution} \\
   $s$/$g$ & $N_c$ & $\left [ -\infty, +\infty \right ]$ & \scriptsize{Prediction scores}  \\
   $P$/$Q$ & $N_c$$\times$$N_c$ & $\left [ 0,1 \right ]$ & \scriptsize{Co-relation matrix}   \\
\hline
\Xhline{2.\arrayrulewidth}
    \end{tabular}
    \label{symbols}
\end{table}

\subsection{Affordance Co-relation Preserving Strategy}
\label{ACP}
There is a co-relation between the categories of affordances, and capturing such co-relation can enhance the network's capability to accurately perceive object affordances. Therefore, the affordance co-relation preserving (ACP) strategy is designed to exploit the co-relation between affordances, as shown in Fig. \ref{pipeline}. First, we feed the feature representations of the two branches ($d_{exo}$ and $d_{ego}$) into a shared fully connected layer to obtain the prediction scores $s$ and $g$. Then, we align the affordance co-relation between the exocentric and egocentric views by calculating the cross-entropy loss \citep{hinton2015distilling} $L_{ACP}$ of the co-relation matrix of the prediction scores of the two branches:
\begin{equation}
\small
    p_j=\frac{exp\left ( s_j/T \right )}{\sum_k^{N_c}exp\left ( s_k/T \right )},\quad q_j=\frac{exp\left ( g_j/T \right )}{\sum_k^{N_c}exp\left ( g_k/T \right )}, \label{acp1}
\end{equation}
\begin{equation}
\small
    P=p^Tp, Q=q^Tq, \label{acp2}
\end{equation}
\begin{equation}
    L_{ACP}=-\sum_j^{N_c}\sum_k^{N_c} P_{jk} log\left ( Q_{jk} \right ). \label{LACP} 
\end{equation}
where $T$ is a hyper-parameter used to control the degree of attention paid to the correlations between negative labels \citep{hinton2015distilling}. $P_{jk}$ and $Q_{jk}$ denote the correlation between categories $j$ and $k$ in the prediction results. Finally, the total loss can be calculated as:
\begin{equation}
\small
    L=\lambda_1 L_{cls}+\lambda_2 L_{ACP} + \lambda_3 L_{KT}. \label{totalL}
\end{equation}
The hyper-parameters $\lambda_1$, $\lambda_2$ and $\lambda_3$ are utilized to balance the losses $L_{cls}$, $L_{ACP}$ and $L_{KT}$. $L_{cls}$ represents the total cross-entropy losses associated with the classification outcomes generated by both branches.

\begin{figure*}[t]
    \centering
    \begin{overpic}[width=0.965\linewidth]{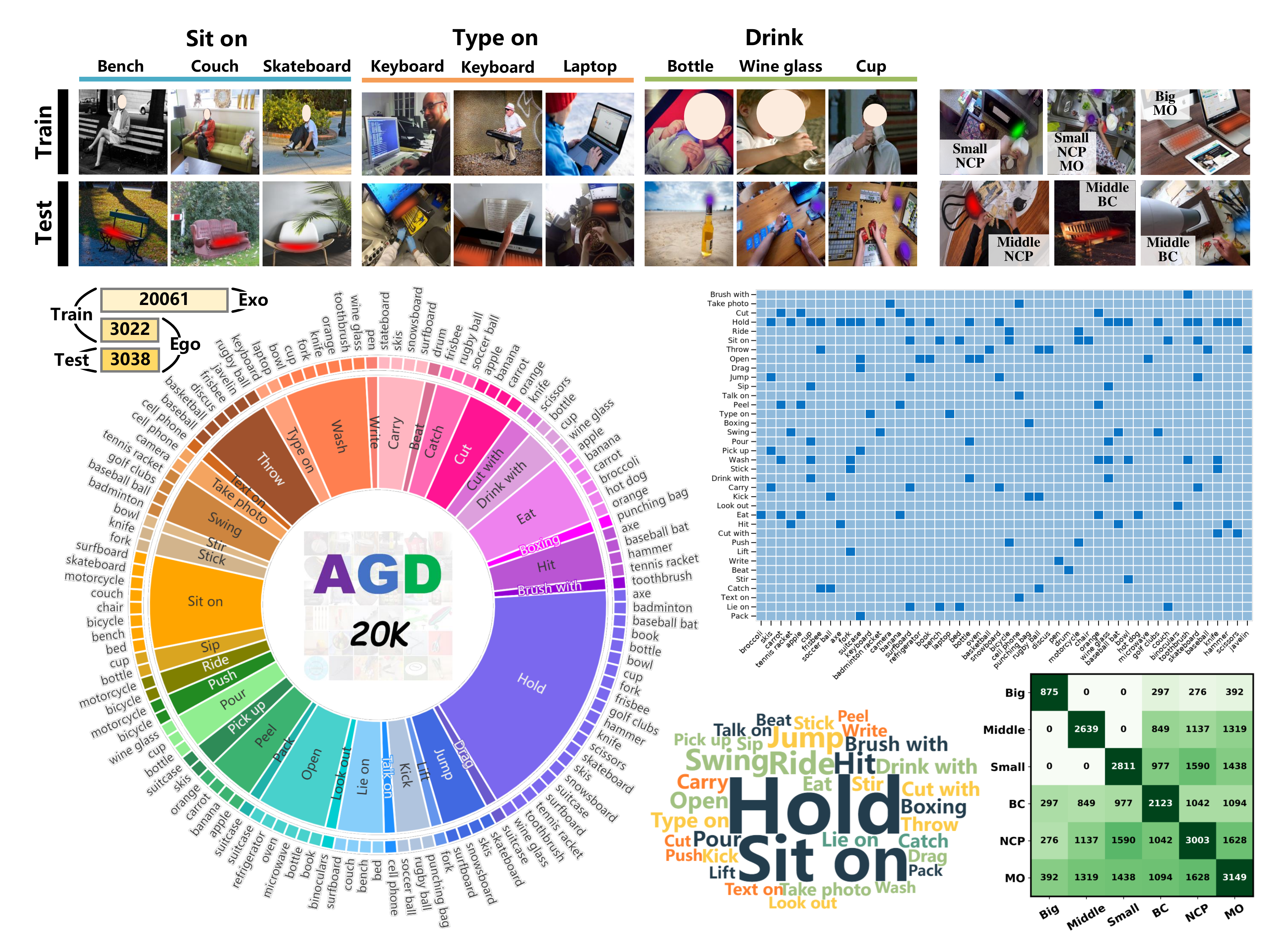}
    \put(1.5,69.5){\colorbox{black}{{\color{white} \textbf{(a)}}}}
    \put(73.8,69.5){\colorbox{black}{{\color{white} \textbf{(c)}}}}
    \put(2,3){\colorbox{black}{{\color{white} \textbf{(b)}}}}
    \put(49.5,50.2){\colorbox{black}{{\color{white} \textbf{(d)}}}}
    \put(50,3){\colorbox{black}{{\color{white} \textbf{(e)}}}}
    \put(74.5,3){\colorbox{black}{{\color{white} \textbf{(f)}}}}
    \end{overpic}
    \vspace{-3pt}
    	\caption{\textbf{Some examples and properties of AGD20K.} \textbf{(a)} Some examples of training and test sets in AGD20K.  \textbf{(b)} The distribution of categories in AGD20K.
      \textbf{(c)} Examples of the different attributes in the AGD20K test set.
     \textbf{(d)} The confusion matrix between the affordance category and the object category in AGD20K, where the horizontal axis denotes the object category, and the vertical axis denotes the affordance category. \textbf{(e)} The word cloud distribution of affordances in AGD20K. \textbf{(f)} The distribution of co-occurring attributes of the AGD20K test set.}
    	\label{FIG:dataset_example}
\end{figure*}

\subsection{Inference}
\label{Inference}
During the testing process, given only an egocentric image and an affordance label, the network localizes the corresponding affordance region (as shown in Fig. \ref{pipeline}). Specifically, we utilize the class activation mapping (CAM) \citep{zhou2016learning} by computing a weighted sum of the feature maps $D_{ego}^i$ of the last convolutional: 
$Y^{C_a}=\sum_i w^{C_a}_i D_{ego}^i$,
where $C_a$ is the affordance class, $D_{ego}^i$ is the $i$-th layer feature map of $D_{ego}$, and $w^{C_a}_i$ is the weight of the fully connected layer corresponding to the $i$-th neuron and the $C_a$ category.

\section{Dataset}
\label{sec:dataset}
\subsection{Dataset Collection} 
\label{dataset collection}
Based on the interactions that often occur in human daily life and the commonly used objects, we select $36$ affordance categories, including indoor and outdoor scenarios in different weather conditions. The exocentric imagery is sourced primarily from COCO \citep{lin2014microsoft} and HICO \citep{chao2018learning}, supplemented by data from PAD \citep{luo2021one}, OPRA \citep{fang2018demo2vec}, and UCF101 \citep{soomro2012ucf101}. Images from the HICO, COCO, and PAD are sorted based on affordance categories, then manually filtered to remove images exhibiting ambiguous interactions. The datasets mentioned above lack adequate examples of images for some interactions, and some affordance categories have no corresponding examples. We expand the dataset by selecting appropriate images from the OPRA \citep{fang2018demo2vec} and UCF101 \citep{soomro2012ucf101} datasets to address this issue. We select video frames depicting human-object interactions with extended durations from various videos, including numerous diverse and intricate examples within the same interaction process. To enrich the diversity of the dataset, we download and filter $2,112$ exocentric images from free-license websites according to interaction and object categories.  
\par There are two main components for the egocentric image: (1) Perception of how objects interact when the intelligent agent enters a new scene. In this instance, the egocentric image does not contain human-object interactions. (2) The perception of the object being interacted with and the interactable region of the surrounding objects when the intelligent body interacts with the object, in which case the egocentric image contains the human-object interaction. We download and select $4,744$ images from the free-license websites for the first case. For the second case, we collect $1,316$ images from EPIC-KITCHENS \citep{damen2018scaling}, Ego4dD \citep{grauman2022ego4d}, THU-READ \citep{tang2017action} \etc, covering a wide range of complex human-object interactions in different environments. Compared to the original dataset, we add $989$ and $1,316$ images for the first and second cases, respectively. Thus, the number of images for both cases in the test set is approximately equivalent. These two components represent the complete egocentric views within the AGD20K, encompassing the agent's perception and interaction with its environment. Examples are shown in Fig. \ref{FIG:dataset_example} (a).

\subsection{Dataset Annotation} 
\label{dataset ann}
For each exocentric image, we assign affordance and object category labels based on the human-object interaction observed in the image. Given the object class contained in each affordance class, we assign affordance labels based on the object class in the egocentric images. Fig. \ref{FIG:dataset_example} (c) shows the confusion matrix between the affordance and the object categories. We adopt heatmaps as part-level labels in the test set to better describe the ``action possibilities'' (\ie, affordance). Specifically, we refer to the OPRA dataset \citep{fang2018demo2vec} for annotating interaction regions, and the annotation routine from previous visual saliency works \citep{bylinskii2015saliency,bylinskii2018different,judd2012benchmark}. By observing the interactions between humans and objects in the exocentric images, we label the egocentric images with points of different densities according to the probability of interaction between the human and object regions. In generating the mask, we apply a Gaussian blur to each labeled point and normalize it to obtain the affordance heatmaps. Some examples are shown in Fig. \ref{FIG:dataset_example} (a). 

\subsection{Statistic Analysis} 
\label{statistic}
To obtain deeper insights into our AGD20K dataset, we show its essential features from the following aspects. The distribution of categories in the dataset is shown in Fig. \ref{FIG:dataset_example} (b), which shows that the dataset contains a wide range of affordance/object categories in diverse scenarios. The affordance and object categories confusion matrix is shown in Fig. \ref{FIG:dataset_example} (d). It shows a multi-to-multi relationship between affordance and object categories, posing a significant challenge for the affordance grounding task. Fig. \ref{FIG:dataset_example} (e) shows the word cloud statistics of AGD20K, implying an unbalanced data distribution, which also satisfies the fact that different interactions occur at different frequencies in the real world scenario. We divide the test set into three subsets, ``\textbf{Big}'', ``\textbf{Middle}'', and ``\textbf{Small}'', according to the scale of the affordance region, \ie, ``Big'': if the proportion of the mask to the whole image is greater than $0.1$, ``Middle'': if the ratio is between $0.03$ and $0.1$, and ``Small'' for the remaining data. Furthermore, we split the test set into three subsets, ``\textbf{BC}'' (\textbf{B}ackground \textbf{C}lutter) \citep{swain1991color}, ``\textbf{NCP}'' (\textbf{N}egative \textbf{C}entral \textbf{P}osition) \citep{lv2022towards} and ``\textbf{MO}'' (\textbf{M}ultiple \textbf{O}bjects), according to the image background complexity, distance from the center, and whether it contains multiple objects. Fig. \ref{FIG:dataset_example} (C) shows some examples of the attributes, while Fig. \ref{FIG:dataset_example} (f) illustrates the confusion matrix for correlating different attributes in the test set. Note that one image may have multiple attributes, increasing the difficulty of locating affordance regions.

\begin{table*}[!t]
    \centering
  \footnotesize
  \renewcommand{\arraystretch}{1.}
  \renewcommand{\tabcolsep}{7.2 pt}
   \caption{\textbf{The results of different models on the original/additional test set.} We compare the results of eight models, DeepGazeII \citep{kummerer2016deepgaze}, EgoGaze , EIL \citep{mai2020erasing}, SPA \citep{pan2021unveiling}, TS-CAM \citep{gao2021ts}, BAS \citep{BAS}, Hotspots \citep{nagarajan2019grounded} and Cross-view-AG \citep{luo2022learning}, on the original test set and the newly added test set. The score before the slash represents the results of the original test set while the score after the slash represents the results of the additional test set. The best results are in \textbf{bold}. }
   \label{Table:1}
  \begin{tabular}{r||ccc|ccc}
    \hline
    \Xhline{2.\arrayrulewidth}
   \multirow{2}{*}{\textbf{Method}} & \multicolumn{3}{c|}{\textbf{Seen}} & \multicolumn{3}{c}{\textbf{Unseen}}   \\
   \cline{2-7}
    & $\text{KLD} \downarrow$ & $\text{SIM} \uparrow$ & $\text{NSS} \uparrow$  & $\text{KLD} \downarrow$ & $\text{SIM} \uparrow$ & $\text{NSS} \uparrow$
\\   \hline
\Xhline{2.\arrayrulewidth}
  DeepGazeII  & $1.858$ / $1.910$  & $0.280$ / $0.259$  & $0.623$ /  $0.678$ & $1.990$ / $2.032$  & $0.256$ / $0.243$ & $0.597$ /  $0.707$ \\
  EgoGaze & $4.185$ / $4.194$  & $0.227$ / $0.222$  & $0.333$ / $0.438$  & $4.285$ / $4.537$  & $0.211$ / $0.193$  & $0.350$ / $0.401$ \\
  \hline   
  EIL  & $1.931$ / $1.903$  & $0.285$ / $0.274$  &  $0.522$ / $0.778$  & $2.167$ / $2.141$  & $0.227$ / $0.226$ & $0.330$ / $0.577$   \\
   SPA  & $5.528$ / $5.779$    & $0.221$ / $0.232$  & $0.357$ / $0.506$ & $7.425$ /  $7.376$  & $0.169$ / $0.193$ & $0.262$ / $0.390$ \\
 TS-CAM & $1.842$ / $1.930$   & $0.260$ / $0.238$  & $0.336$ /  $0.496$  & $2.104$ / $2.176$  & $0.201$ / $0.196$   & $0.151$ / $0.267$    \\
  BAS  & $1.925$ / $1.951$   & $0.279$ / $0.241$  & $0.702$ / $0.763$ & $2.216$ / $2.226$ & $0.226$ / $0.208$   & $0.531$ / $0.570$   \\
  \hline
  Hotspots  & $1.773$ / $2.136$    & $0.278$ / $0.208$  & $0.615$ / $0.368$  & $1.994$ / $2.377$  & $0.237$ / $0.179$  & $0.577$ / $0.243$ \\
  \hline 
  Cross-view-AG  & $1.538$ / $1.684$ & $0.334$ / $0.296$  & $0.927$ / $1.071$  & $1.787$ / $1.936$ & $\bm{0.285}$ / $0.249$  & $0.829$ /  $0.905$ \\
  \rowcolor{mygray}
  \textbf{Ours} & $\bm{1.478}$ / $\bm{1.576}$  & $\bm{0.342}$ / $\bm{0.314}$  & $\bm{1.012
  }$ / $\bm{1.228}$ &    $\bm{1.749}$  / $\bm{1.848}$ &  $0.281$ / $\bm{0.258}$  & $\bm{0.897}$ / $\bm{1.050}$   \\
    \hline
    \Xhline{2.\arrayrulewidth}
    \end{tabular}
  \end{table*}

\begin{table*}[!t]
    \centering
  \footnotesize
  \renewcommand{\arraystretch}{1.}
  \renewcommand{\tabcolsep}{9.1 pt}
   \caption{\textbf{The results of different methods on the mixed testset.} The best results are in \textbf{bold}. ``Seen'' means that the training set and the test set contain the same object categories, while ``Unseen'' means that the object categories in the training set and the test set do not overlap. The $\textcolor{darkpink}{\diamond}$ defines the relative improvement of our method over other methods.}
   \label{Table:mix dataset}
  \begin{tabular}{r||ccc|ccc}
    \hline
    \Xhline{2.\arrayrulewidth}
  \multirow{2}{*}{\textbf{Method}} & \multicolumn{3}{c|}{\textbf{Seen}} & \multicolumn{3}{c}{\textbf{Unseen}}   \\
  \cline{2-7}
    & $\text{KLD} \downarrow$ & $\text{SIM} \uparrow$ & $\text{NSS} \uparrow$  & $\text{KLD} \downarrow$ & $\text{SIM} \uparrow$ & $\text{NSS} \uparrow$
\\   \hline
\Xhline{2.\arrayrulewidth}
  DeepGazeII  & $1.899\textcolor{darkpink}{\scriptstyle~\diamond18.1\%}$ &	$0.264\textcolor{darkpink}{\scriptstyle~\diamond22.0\%}$	& $0.663\textcolor{darkpink}{\scriptstyle~\diamond76.5\%}$ &	$2.020\textcolor{darkpink}{\scriptstyle~\diamond9.3\%}$ &	$0.246\textcolor{darkpink}{\scriptstyle~\diamond6.1\%}$ &	$0.677\textcolor{darkpink}{\scriptstyle~\diamond50.2\%}$  \\
  EgoGaze  &  $4.195\textcolor{darkpink}{\scriptstyle~\diamond62.9\%}$ &	$0.223\textcolor{darkpink}{\scriptstyle~\diamond44.4\%}$ &	$0.409\textcolor{darkpink}{\scriptstyle~\diamond186.1\%}$ &	$4.467\textcolor{darkpink}{\scriptstyle~\diamond59.0\%}$	& $0.198\textcolor{darkpink}{\scriptstyle~\diamond31.8\%}$ &	$0.387\textcolor{darkpink}{\scriptstyle~\diamond162.8\%}$ \\
  \hline    
  EIL  &  $1.914\textcolor{darkpink}{\scriptstyle~\diamond18.8\%}$ &	$0.276\textcolor{darkpink}{\scriptstyle~\diamond16.7\%}$ &	$0.708\textcolor{darkpink}{\scriptstyle~\diamond65.3\%}$ & $2.148\textcolor{darkpink}{\scriptstyle~\diamond14.7\%}$ & $0.226\textcolor{darkpink}{\scriptstyle~\diamond15.5\%}$ & $0.509\textcolor{darkpink}{\scriptstyle~\diamond99.8\%}$ \\
  SPA &  $5.719\textcolor{darkpink}{\scriptstyle~\diamond72.8\%}$ & $0.229\textcolor{darkpink}{\scriptstyle~\diamond40.6\%}$ &	$0.466\textcolor{darkpink}{\scriptstyle~\diamond151.1\%}$ &	$7.399\textcolor{darkpink}{\scriptstyle~\diamond75.2\%}$ &	$0.186\textcolor{darkpink}{\scriptstyle~\diamond40.3\%}$ &	$0.351\textcolor{darkpink}{\scriptstyle~\diamond189.7\%}$ \\
  TS-CAM &  $1.909\textcolor{darkpink}{\scriptstyle~\diamond18.5\%}$ &	$0.243\textcolor{darkpink}{\scriptstyle~\diamond32.5\%}$ &	$0.453\textcolor{darkpink}{\scriptstyle~\diamond158.3\%}$ &	$2.129\textcolor{darkpink}{\scriptstyle~\diamond14.0\%}$ &	$0.205\textcolor{darkpink}{\scriptstyle~\diamond27.3\%}$ &	$0.277\textcolor{darkpink}{\scriptstyle~\diamond267.1\%}$   \\
  BAS & $1.945\textcolor{darkpink}{\scriptstyle~\diamond20.1\%}$ &	$0.251\textcolor{darkpink}{\scriptstyle~\diamond28.3\%}$ &	$0.748\textcolor{darkpink}{\scriptstyle~\diamond56.4\%}$	&	$2.223\textcolor{darkpink}{\scriptstyle~\diamond17.6\%}$ &	$0.213\textcolor{darkpink}{\scriptstyle~\diamond22.5\%}$ &	$0.560\textcolor{darkpink}{\scriptstyle~\diamond81.6\%}$   \\
  \hline
  Hotspots &  $2.104\textcolor{darkpink}{\scriptstyle~\diamond26.1\%}$ &	$0.215\textcolor{darkpink}{\scriptstyle~\diamond49.8\%}$ &	$0.356\textcolor{darkpink}{\scriptstyle~\diamond228.7\%}$ &	$2.332\textcolor{darkpink}{\scriptstyle~\diamond21.4\%}$ &	$0.184\textcolor{darkpink}{\scriptstyle~\diamond41.8\%}$ &	$0.245\textcolor{darkpink}{\scriptstyle~\diamond315.1\%}$  \\
  \hline 
  Cross-view-AG  &  $1.647\textcolor{darkpink}{\scriptstyle~\diamond5.6\%}$ &	$0.306\textcolor{darkpink}{\scriptstyle~\diamond5.3\%}$ &	$1.032\textcolor{darkpink}{\scriptstyle~\diamond13.4\%}$ &	$1.895\textcolor{darkpink}{\scriptstyle~\diamond3.3\%}$ &	$0.259\textcolor{darkpink}{\scriptstyle~\diamond0.8\%}$ &	$0.884\textcolor{darkpink}{\scriptstyle~\diamond15.0\%}$  \\
  \rowcolor{mygray}
  \textbf{Ours} & $\bm{1.555}_{\pm 0.007}$ &	$\bm{0.322}_{\pm 0.001}$ &	$\bm{1.170}_{\pm 0.020}$ &	$\bm{1.832}_{\pm 0.005}$ &	$\bm{0.261}_{\pm 0.003}$ &	$\bm{1.017}_{\pm 0.012}$ \\
    \hline
    \Xhline{2.\arrayrulewidth}
    \end{tabular}
  \end{table*}
  
\begin{table*}[t]
  \renewcommand{\arraystretch}{1.}
\renewcommand{\tabcolsep}{7pt}
  \caption{\textbf{Parameters (M) and inference time (s) for all models.} }
  \centering
 \small
  \begin{tabular}{c||cc|cccc|c|c||c}
    \hline
    \Xhline{2.\arrayrulewidth}
     \textbf{Method} &  DeepGazeII   & EgoGaze  & EIL  & SPA   & TS-CAM  & BAS  & Hotspots   & Cross-view-AG   & \textbf{Ours}  \\
    \hline
    \Xhline{2.\arrayrulewidth}
    \textbf{Param. (M)} &  $20.44$ & $46.53$ & $42.41$ & $69.28$ & $85.86$ & $53.87$ & $132.64$  &  $120.03$ & $82.27$ \\
    \textbf{Time (s)} &  $3.760$ & $0.026$ & $0.019$ & $0.081$ & $0.023$ & $0.057$ & $0.087$ & $0.023$ &  $0.022$ \\
    \hline
    \Xhline{2.\arrayrulewidth}
    \end{tabular}
  \label{Table:FLOPS}
\end{table*}

\section{Experiments}
\label{sec:experiments}
\subsection{Metrics}
\label{Metrics}
Previous works mainly segment precise affordance regions \citep{luo2021one,chuang2018learning,nguyen2017object,myers2015affordance}, while the cross-view affordance grounding task considers a weakly supervised setting that predicts the affordance heatmap using only the affordance category label. Referring to the hotspots grounding-related works \citep{fang2018demo2vec,nagarajan2019grounded,liu2022joint,bylinskii2018different}, we adopt heatmaps to give a better description of the ``action possibilities'' (\ie, affordance) and use \textbf{KLD}  \citep{bylinskii2018different}, \textbf{SIM} \citep{swain1991color}, and \textbf{NSS} \citep{peters2005components} to evaluate the probability distribution correlation between the predicted affordance heatmap and Ground Truth (GT).

\begin{itemize}
\item [\textbf{-}]  \textbf{K}ullback-\textbf{L}eibler \textbf{D}ivergence (\textbf{KLD}) \citep{bylinskii2015saliency} measures the distribution difference between the prediction ($P$) and the ground truth ($Q$). It is computed as follows:
\begin{equation}
\small
   KLD\left ( P,Q^{D} \right )=\sum_{i}Q_{i}^{D}log\left ( \epsilon + \frac{Q_{i}^{D}}{\epsilon+P_{i}} \right ), \label{eq:no20}
\end{equation}
where $\epsilon$ is a regularization constant.

\item [\textbf{-}] \textbf{Sim}ilarity (\textbf{SIM}) \citep{swain1991color}  measures the similarity between the prediction map ($P$) and the continuous ground truth map ($Q^{D}$). It is computed as follows:
\begin{equation}
   SIM\left ( P, Q^{D} \right )=\sum_{i}min\left ( P_{i},Q_{i}^{D}\right ),\\
\end{equation}
where  $\sum_{i}P_{i}=\sum_{i}Q_{i}^{D}=1$.
\item [\textbf{-}] \textbf{N}ormalized \textbf{S}canpath \textbf{S}aliency (\textbf{NSS}) \citep{peters2005components} measures the correspondence between the prediction map ($P$) and the ground truth ($Q^{D}$). It is computed as follows:
\begin{equation}
\small
   NSS\left ( P,Q^{D} \right )=\frac{1}{N}\sum_{i}\hat{P}\times Q_{i}^{D}, \label{eq:no22}
\end{equation}
where $N=\sum_{i}Q_{i}^{D}$, $\hat{P}=\frac{P-\mu\left ( P \right )}{\sigma\left ( P \right )}$. $\mu\left ( P \right )$ and $\sigma\left ( P \right )$ are the mean and standard deviation, respectively.
\end{itemize}

\begin{figure*}[t]
    \centering
    \begin{overpic}[width=1.\linewidth]{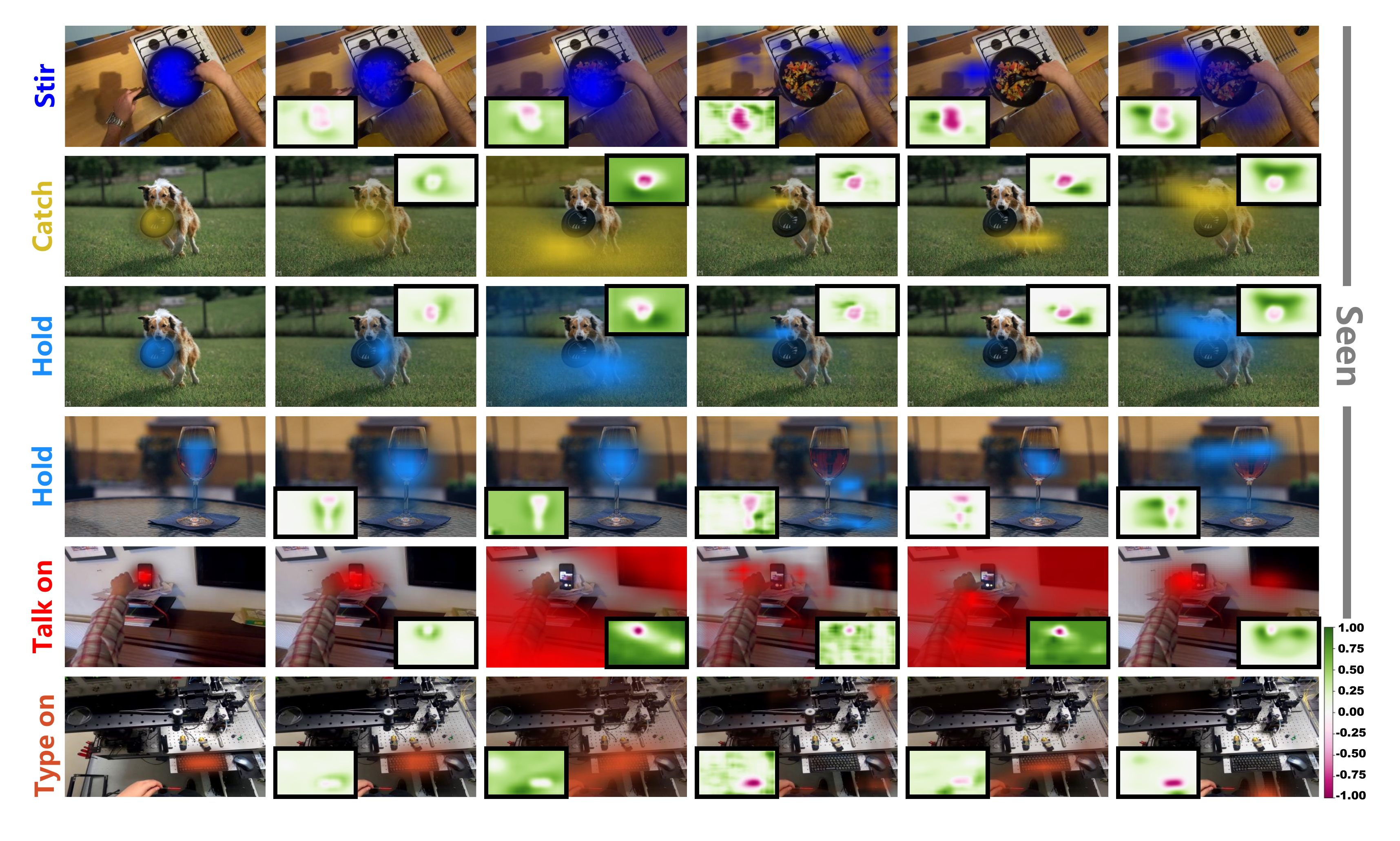}
    \put(8.3,0.6){\colorbox{white}{{\color{black} \textbf{GT}}}}
    \put(23.5,0.6){\colorbox{white}{{\color{black} \textbf{Ours}}}}
    \put(33.5,0.6){\colorbox{white}{{\textbf{Cross-view-AG}}}}
   
    \put(51.9,0.6){\colorbox{white}{{\textbf{Hotspots}}}}
    \put(69.4,0.6){\colorbox{white}{{\textbf{EIL}}}}
     \put(80,0.6){\colorbox{white}{{\textbf{DeepGazeII}}}}
    \end{overpic}
    \caption{\textbf{ Visualization of the difference maps at the ``Seen'' setting.} We select results from representative models, such as Cross-view-AG \citep{luo2022learning}, Hotspots \citep{nagarajan2019grounded}, EIL \citep{mai2020erasing} and DeepGazeII \citep{kummerer2016deepgaze} for presentation. The difference maps represent the difference between the prediction and the GT. }
    \label{figure:main_result_seen}
\end{figure*} 

\begin{figure*}[t]
    \centering
   \begin{overpic}[width=1.\linewidth]{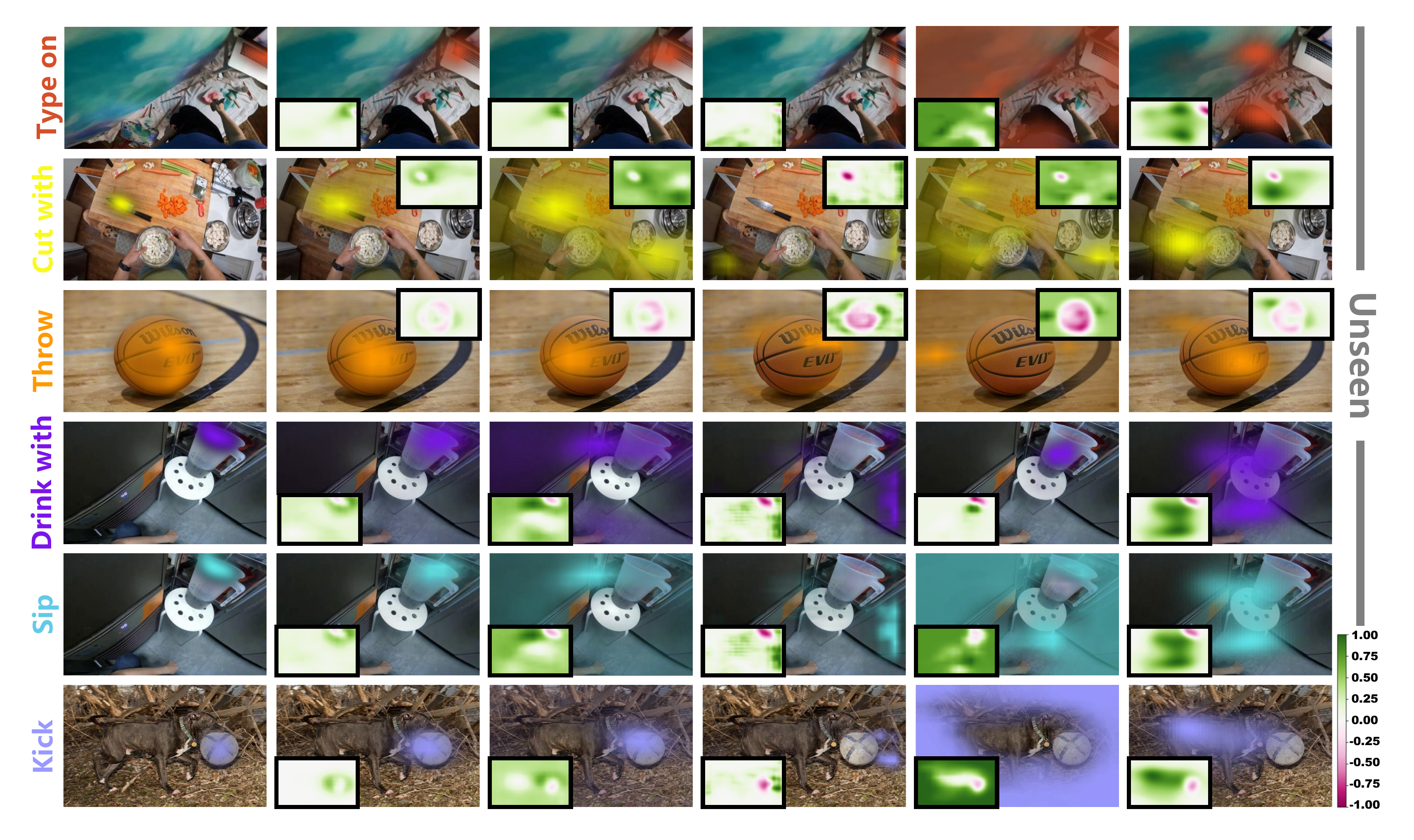}
    \put(8.3,0.1){\colorbox{white}{{\color{black} \textbf{GT}}}}
    \put(23.5,0.1){\colorbox{white}{{\color{black} \textbf{Ours}}}}
    \put(33.5,0.1){\colorbox{white}{{\textbf{Cross-view-AG}}}}
   
    \put(51.9,0.1){\colorbox{white}{{\textbf{Hotspots}}}}
    \put(69.4,0.1){\colorbox{white}{{\textbf{EIL}}}}
     \put(80,0.1){\colorbox{white}{{\textbf{DeepGazeII}}}}
    \end{overpic}
    \caption{\textbf{Visual affordance heatmaps and difference maps at the ``Unseen'' setting.} We select results from representative models, such as Cross-view-AG \citep{luo2022learning}, Hotspots \citep{nagarajan2019grounded}, EIL \citep{mai2020erasing} and DeepGazeII \citep{kummerer2016deepgaze} for presentation. The difference maps represent the difference between the prediction and the GT. }
    \label{figure:main_result_unseen}
\end{figure*}

\subsection{Comparison Methods}
\label{Comparison Methods}
To comprehensively evaluate the effectiveness of our approach in cross-view affordance grounding tasks, we choose eight advanced methods in four relevant fields, such as saliency detection (\textbf{DeepGazeII \citep{kummerer2016deepgaze}}, \textbf{EgoGaze \citep{huang2018predicting}}), weakly supervised object localization (WSOL) (\textbf{EIL} \citep{mai2020erasing}, \textbf{SPA}  \citep{pan2021unveiling}, \textbf{TS-CAM} \citep{gao2021ts}, \textbf{BAS}  \citep{BAS}), affordance grounding (\textbf{Hotspots} \citep{nagarajan2019grounded}) and Cross-view affordnace knowledge transfer (\textbf{Cross-view-AG}  \citep{luo2022learning}) for comparison.  For the saliency detection models, we use models trained on the saliency datasets and test in the same way as \citep{nagarajan2019grounded}.  For the weakly supervised object localization models, we only utilize exocentric images for training.
\begin{itemize}
\item [\textbf{-}] \textbf{DeepGazeII} \citep{kummerer2016deepgaze}: Unlike other saliency models, it does not perform additional fine-tuning of the VGG features and only trains some output layers to predict saliency on top of VGG \citep{simonyan2014very}.
\item [\textbf{-}] \textbf{EgoGaze} \citep{huang2018predicting}: The proposed model is a hybrid approach that merges task-dependent attention transitions and bottom-up saliency prediction to generate gaze predictions.
\item [\textbf{-}] \textbf{EIL} \citep{mai2020erasing}:  It introduces a novel adversarial erasing technique jointly exploring highly response class-specific areas and less discriminative regions to obtain a complete object region.
\item [\textbf{-}] \textbf{SPA} \citep{pan2021unveiling}: It explores how to extract object structure information during training and proposes a structure-preserving activation method that leverages the structure information incorporated in the convolutional features for WSOL task.
\item [\textbf{-}] \textbf{TS-CAM} \citep{gao2021ts}: It proposes a token semantic coupled attention map to take full advantage of the self-attention mechanism in visual transformer for long-range
dependency extraction. 
\item [\textbf{-}] \textbf{BAS} \citep{BAS}: The proposed model enhances the accuracy of foreground map generation through an activation mapping constraint module. This module helps in learning predicted maps by restraining background activation, resulting in more precise predictions.
\item [\textbf{-}] 
\textbf{Hotspots} \citep{nagarajan2019grounded}: It is a weakly supervised way to learn the affordance of an object through video, and affordance grounding is achieved only through action labels.

\item [\textbf{-}] \textbf{Cross-view-AG} \citep{luo2022learning}:  The model extracts affordance invariance cues from diverse exocentric interactions and transfer it to egocentric view. Furthermore, It improves the perception of affordance regions through the preservation of affordance co-relation
\end{itemize}

\subsection{Implementation Details}
\label{Implementation Details}
Our model is implemented in PyTorch and trained with the SGD optimizer. With random horizontal flipping, the input images are randomly cropped from 256$\times$256 to 224$\times$224. We train the model for 35 epochs on a single NVIDIA 3090ti GPU with an initial learning rate of 1$e$-3. The hyper-parameters $\lambda_1$, $\lambda_2$, and $\lambda_3$ are set to 1, 0.5, and 0.5, respectively. We set the batch size to 32, and the number of exocentric images $N$ is set to 3. The hyper-parameter $T$ in the ACP strategy is set to 1. The dictionary matrix $W$'s rank $r$ and the number of iterations in the AIM module are set to 64 and 6, respectively. The number of channels of input features in the AIM module is 64. We divide the dataset into ``Seen'' and ``Unseen'', in which ``Seen'' means that the training and test sets contain the same class of objects, while ``Unseen'' indicates that the training and test sets contain different classes of objects. The ``Unseen'' split can be used to evaluate the generalization ability of the models. We use Resnet50 \citep{he2016deep} as the backbone while other advanced backbones~\citep{liu2021swin,zhang2023vitaev2} can be explored in future work. We use the same data augmentation method and batch size for the other methods. For weakly supervised object localization models (EIL \citep{mai2020erasing}, SPA \citep{pan2021unveiling}, TS-CAM \citep{gao2021ts} and BAS \citep{BAS}), the input is exocentric images during training, while the test input is egocentric images. For Hotspots \citep{nagarajan2019grounded}, three images are randomly sampled from the exocentric images during training and used as input to the video branch.

\subsection{Quantitative and Qualitative Comparisons}
\label{Quantitative and Qualitative Comparisons}
Table \ref{Table:1} shows the performance of the different models on the original test set and the newly supplemented test set. The scores before and after the slash denote the outcomes on the original and newly established test sets. For models such as DeepGazeII \citep{kummerer2016deepgaze}, BAS \citep{BAS}, and Cross-view-AG \citep{luo2022learning}, which perform better in the original test set, they all show a performance drop on the new test set in most metrics. EIL \citep{mai2020erasing} explores the whole object through adversarial erasure techniques.  Thus there is no particular impact on performance for attributes such as complex backgrounds. However, it is not easy to obtain accurate part-level localization. Compared to the previous state-of-the-art Cross-view-AG \citep{luo2022learning} framework, the performance of our model degrades less on the new test set.  Moreover, the evidence is more obvious in the ``Unseen'' setting (see in Table \ref{Table:1} right), demonstrating the better generalization ability of our model in handling complex scenarios.
\par The subsequent experiments and the corresponding analyses are conducted on a combined dataset consisting of a mixture of the original and the new test sets. Table \ref{Table:mix dataset} shows the results for different related models, and it is evident that our approach achieves the best results for all metrics on both the Seen and Unseen settings. Taking KLD as the metric, our method improves \textbf{18.1\%} compared to the best saliency model, \textbf{18.5\%} over the best weakly supervised object localization (WSOL) model, \textbf{26.1\%} over the affordance grounding model, and \textbf{5.6\%} over the best cross-view affordance knowledge transfer model in the ``Seen'' setting. Our method on the ``Unseen'' setting improves \textbf{9.3\%} compared to the best saliency model, surpasses the best WSOL model by \textbf{14.0\%},  exceeds the affordance grounding model by \textbf{21.4\%}, and outperforms the cross-view affordance knowledge transfer model by \textbf{3.3\%}. Table \ref{Table:FLOPS} presents the various models' parameter numbers and inference times. The number of parameters in our model is noticeably lower than Hotspots \citep{nagarajan2019grounded} and Cross-view-AG \citep{luo2022learning}, and comparable to TS-CAM \citep{gao2021ts}. However, our model still has a larger number of parameters than the other methods, and thus, future work should aim to further reduce its size. Furthermore, the inference time of our model is comparable to that of various other methods.

Fig. \ref{figure:main_result_seen} and Fig. \ref{figure:main_result_unseen} present the affordance maps for the ``Seen'' and ``Unseen'' settings, along with the discrepancy maps between the predicted outcomes and the actual ground truth. Compared to other models, our method can more accurately locate the affordance region of the object. Specifically, the red areas in the difference map are generally small and light, which indicates that our method can generate a complete affordance region of the object. For images with complex backgrounds (\eg, Fig. \ref{figure:main_result_seen}, row 5 and Fig. \ref{figure:main_result_unseen}, row 6), the model also can locate the affordance regions more accurately, indicating that our method generalizes well in complex scenes. For the cases where an object belongs to more than one affordance class or the same affordance contains multiple object classes with vastly different appearances, the model can accurately find the corresponding region, demonstrating that our method can effectively address the challenges posed by multiple possibilities of affordances.

\begin{table}[!t]
    \centering
 \footnotesize
  \renewcommand{\arraystretch}{1.}
  \renewcommand{\tabcolsep}{2.15 pt}
   \caption{\textbf{Long Tail Distribution.} We divide AGD20K into two subsets (``Head'' and ``Tail'', according to the number of images in the affordance class), and test the performance of the models in the two subsets separately.}
   \vspace{-2pt}
   \label{Table:long tail}
  
  \begin{tabular}{r||ccc|ccc}
  \Xhline{2.\arrayrulewidth}
    \hline
 \multirow{3}{*}{\textbf{Method}} & \multicolumn{6}{c}{\begin{minipage}[b]{0.6\columnwidth}
		\centering
		\raisebox{-.1\height}{\includegraphics[width=0.98\linewidth]{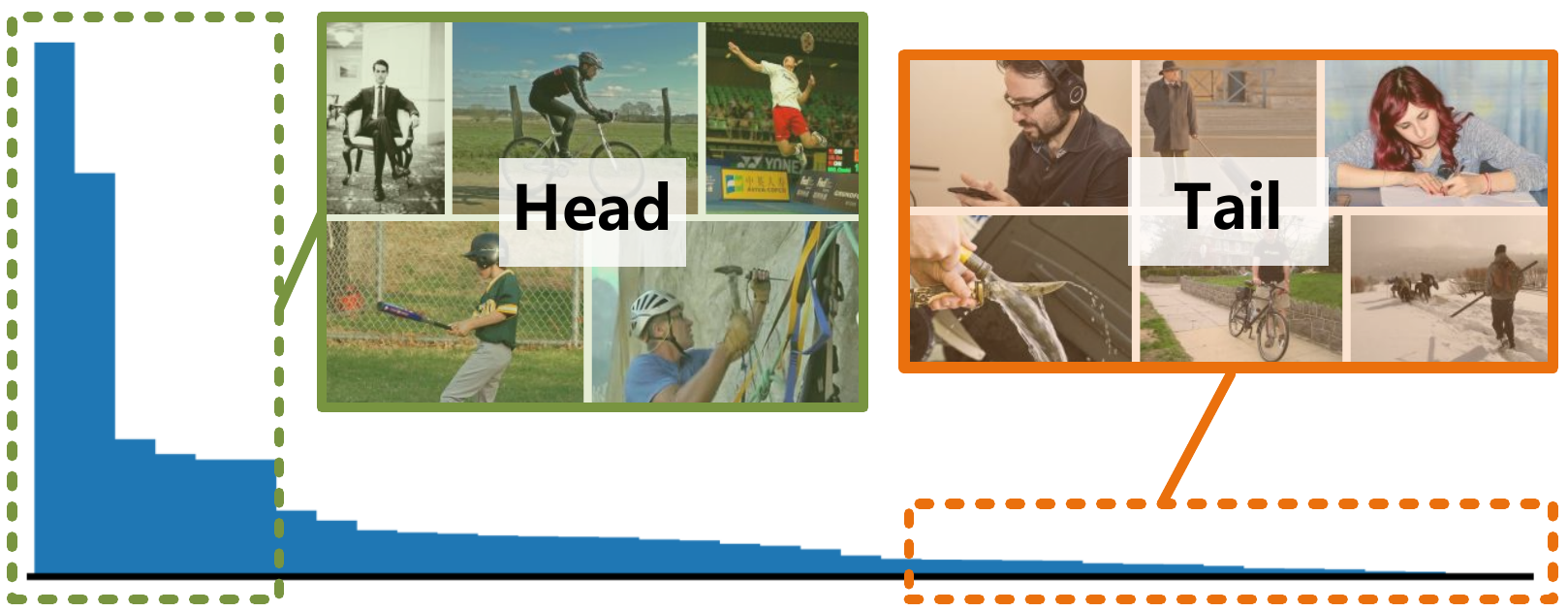}}
	\end{minipage}}  \\
	\cline{2-7}
 & \multicolumn{3}{c|}{\textbf{Head}} & \multicolumn{3}{c}{\textbf{Tail}}  \\
  \cline{2-7}
     & {$\text{KLD}\downarrow$} & $\text{SIM}\uparrow$ & $\text{NSS}\uparrow$ & $\text{KLD}\downarrow$ & $\text{SIM}\uparrow$ & $\text{NSS}\uparrow$  \\
   \hline
   \Xhline{2.\arrayrulewidth}
   DeepGazeII  & $1.921$ & $0.250$ & $0.636$ & $1.925$ & $0.269$ & $0.662$  \\
  EgoGaze  & $4.132$ & $0.222$ & $0.414$ & $4.370$ & $0.206$ & $0.366$\\
   EIL   &  $1.980$ & $0.269$ & $0.658$ & $2.036$ & $0.258$ & $0.622$ \\
   SPA  & $4.176$ & $0.226$ & $0.424$ & $7.997$ & $0.205$ & $0.368$  \\
   TS-CAM  & $1.789$ & $0.257$ & $0.698$ & $2.130$ & $0.213$ & $0.126$ \\
   BAS  & $2.210$ & $0.225$ & $0.550$ & $1.988$ & $0.233$ & $0.661$  \\
  Hotspots  &  $2.110$ & $0.209$ & $0.324$ & $2.204$ & $0.205$ &  $0.309$ \\
  Cross-view-AG  &  $1.691$ &	$0.288$ &	$0.931$ &	$1.812$ &	$0.284$ &	$0.928$ \\
   \hline
   \rowcolor{mygray}
   \textbf{Ours} & $\bm{1.607}$ &	$\bm{0.303}$ &	$\bm{1.045}$ &	$\bm{1.713}$ &	$\bm{0.300}$ &	$\bm{1.080}$  \\
    \hline
    \Xhline{2.\arrayrulewidth}
    \end{tabular}
\end{table}

\subsection{Performance Analysis}
\noindent\textbf{Long Tail Distribution.\ } The AGD20K dataset exhibits a long-tailed distribution characterized by significant data imbalance, as presented in Table \ref{Table:long tail}. To validate whether our model can perform better on a small number of samples, we select the ``Head'' and ``Tail'' classes and test them separately. The results are shown in Table \ref{Table:long tail}. Our model outperforms the other methods in the ``Head'' and ``Tail'' subsets. This improvement may stem from implementing the ACP strategy, which maintains the relationship between affordance classes and amplifies the network's ability to recognize classes with minimal data.

\noindent\textbf{Different Sources. }To validate that the superiority of our method is not due to the additional egocentric images, we conducted retraining of the weakly supervised object localization models with both exocentric and egocentric images as input. Table \ref{Table:different sources} shows the results, indicating that using exocentric and egocentric images simultaneously enhances most approaches. However, the benefits of additional samples are limited and do not lead to significant gains. Our approach outperforms all these models, demonstrating that explicit knowledge transfer can efficiently transfer affordance knowledge to egocentric perspectives and attain more precise localization results.

\begin{figure*}[t]
    \centering
    \begin{overpic}[width=0.99\linewidth]{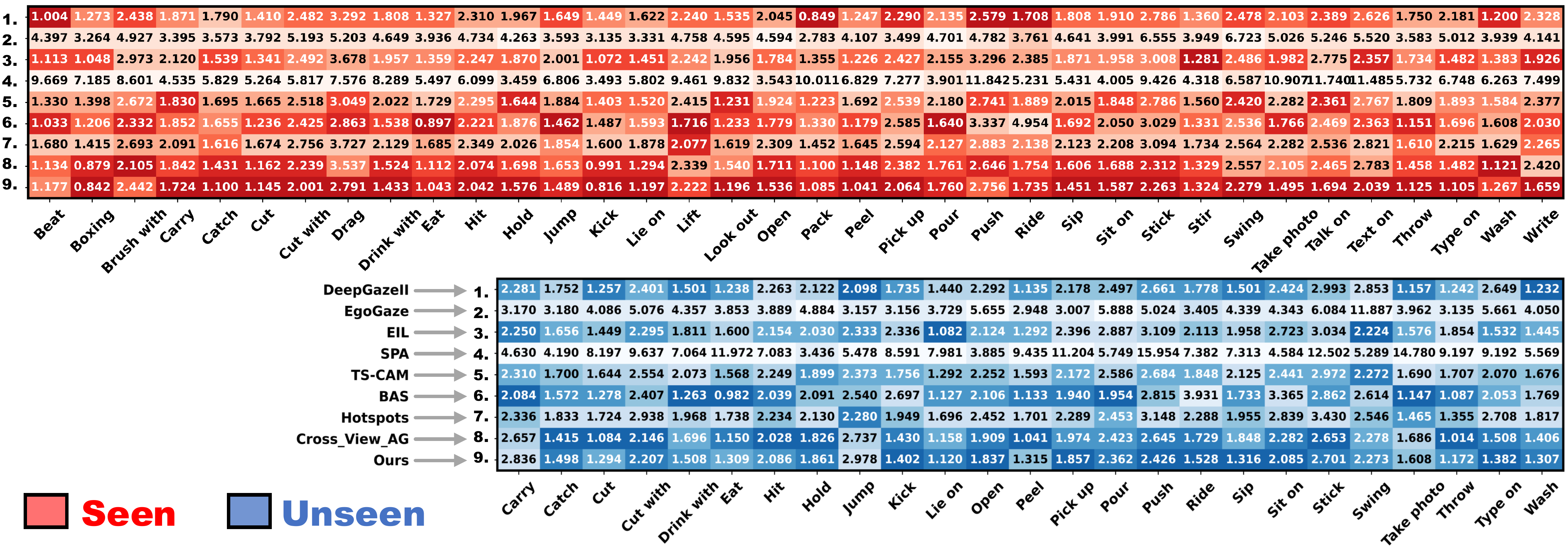}
    \put(4.2,16.1){\tiny\textbf{\citep{kummerer2016deepgaze}}}
    \put(8.5,14.7){\tiny\textbf{\citep{huang2018predicting}}}
    \put(13.,13.4){\tiny\textbf{\citep{mai2020erasing}}}
    \put(12.5,12.0){\tiny\textbf{\citep{pan2021unveiling}}}
    \put(10.5,10.7){\tiny\textbf{\citep{gao2021ts}}}
    \put(12.8,9.3){\tiny\textbf{\citep{BAS}}}
    \put(5.7,7.9){\tiny\textbf{\citep{nagarajan2019grounded}}}
    \put(7.2,6.5){\tiny\textbf{\citep{luo2022learning}}}
    \end{overpic}
    \caption{\textbf{The results of the different methods on the AGD20K for each affordance category.} We calculate the KLD metrics for each affordance category in both ``Seen'' and ``Unseen'' settings, with darker colors representing better model performance. }
    \label{figure:class}
\end{figure*}

\noindent\textbf{Different Classes. }Fig. \ref{figure:class} shows the results of the KLD metrics for each category in both ``Seen''  and ``Unseen'' settings, with deeper colors indicating better performance. Our model achieves the best results under most affordance categories, demonstrating our method's superiority in locating affordance regions. In the ``Seen'' setting, our model obtains more accurate results for both the affordance categories ``Hold'' and ``Cut with'', where there are some co-relations, demonstrating that our approach can enhance the network's perception of affordance regions by aligning the co-relation of the two views. For affordance categories such as ``Open'' and ``Carry'', where the interaction habits of different humans are quite diverse, our method still exceeds all other models, validating the effectiveness of the AIM module in extracting affordance-specific features for localization. In the ``Unseen'' setting, our model achieves promising results for the categories ``Pick up'', ``Sit on'', \etc, with large variations between the object appearances in the training and test sets, demonstrating that our method has a strong generalization ability to new object categories.

\begin{table}[!t]
  \centering
  \renewcommand{\arraystretch}{1.}
  \renewcommand{\tabcolsep}{1.7pt}
   \footnotesize
   \caption{\textbf{Different sources.} ``Exo'' means simply using exocentric images, while ``Both'' means using both exocentric and egocentric images in training.}
   \label{Table:different sources}
  \begin{tabular}{c|c||ccc|ccc}
    \hline
    \Xhline{2.\arrayrulewidth}
   \multirow{2}{*}{\textbf{Method}} & \multirow{2}{*}{\textbf{Source}} 
     & \multicolumn{3}{c|}{\textbf{Seen}} & \multicolumn{3}{c}{\textbf{Unseen}}  \\
    
    \cline{3-8}
     &  & $\text{KLD} \downarrow$ & $\text{SIM} \uparrow$ & $\text{NSS} \uparrow$ & $\text{KLD} \downarrow$ & $\text{SIM} \uparrow$ & $\text{NSS} \uparrow$  \\
   \hline
   \Xhline{2.\arrayrulewidth}
  
   \multirow{2}{*}{ EIL }  & Exo  & $1.914$ &	$0.276$ &	$0.708$ & $2.148$ &	$0.226$ &	$0.509$   \\
    &  Both & $1.983$ & $0.294$ & $0.849$ & $2.078$ & $0.239$ & $0.662$  \\
 \hline
  
 \multirow{2}{*}{SPA }  & Exo & $5.719$ &	$0.229$ &	$0.466$	&	$7.399$ &	$0.186$ &	$0.351$     \\
    
  &  Both & $5.057$ & $0.248$ & $0.540$ & $6.389$ & $0.216$ & $0.478$  \\
  \hline

  \multirow{2}{*}{TS-CAM}   & Exo & $1.909$ &	$0.243$ &	$0.453$ &	$2.129$ &	$0.205$ &	$0.277$  \\
 &  Both &    $1.882$ & $0.254$ & $0.511$ & $2.124$ & $0.206$ & $0.272$ \\

\hline
\multirow{2}{*}{BAS }  & Exo & $1.945$ &	$0.251$ &	$0.748$	&	$2.223$ &	$0.213$ &	$0.560$ \\
&  Both &  $2.188$ & $0.263$ & $0.734$ & $2.002$ & $0.225$ & $0.826$ \\
 \hline
 \rowcolor{mygray}
  \textbf{Ours} & Both & $\bm{1.555}$ &	$\bm{0.322}$ &	$\bm{1.170}$ &	$\bm{1.832}$ &	$\bm{0.261}$ &	$\bm{1.017}$     \\
    \hline
    \Xhline{2.\arrayrulewidth}
    \end{tabular}
\end{table}

\begin{table}[!t]
    \centering
  \footnotesize
  \renewcommand{\arraystretch}{1.}
  \renewcommand{\tabcolsep}{0.5 pt}
   \caption{\textbf{Ablation study.} We examine the effect of the AIM module, CFT module and ACP strategy on results.}
   \label{Table:ablation study1}

  \begin{tabular}{ccc||ccc|ccc}
  \Xhline{2.\arrayrulewidth}
    \hline
 \multirow{2}{*}{\scriptsize{\textbf{AIM}}} & \multirow{2}{*}{\scriptsize{\textbf{CFT}}} & \multirow{2}{*}{\scriptsize{\textbf{ACP}}} & \multicolumn{3}{c|}{\textbf{Seen}} & \multicolumn{3}{c}{\textbf{Unseen}}  \\
  \cline{4-9}
    &   &  & \bm{$\text{KLD} \downarrow$} & \bm{$\text{SIM} \uparrow$} & \bm{$\text{NSS} \uparrow$} & \bm{$\text{KLD} \downarrow$} & \bm{$\text{SIM} \uparrow$} & \bm{$\text{NSS} \uparrow$}  \\
   \hline
   \Xhline{2.\arrayrulewidth}
   & & &    $1.765$ & $0.289$  & $0.878$ & $2.003$ &	$0.247$ &	$0.725$  \\
   $\checkmark$ & & & $1.680$ &	$0.304$ &	$0.974$  & $1.973$ &	$0.249$ &	$0.765$  \\
     &  &  $\checkmark$ &  $1.659$ &	$0.313$ &	$1.009$ & $1.936$ &	$0.248$ &	$0.828$   \\
   $\checkmark$ & $\checkmark$ &  & $1.602$ &	$0.320$ &	$1.089$ & $1.896$ &	$0.256$ &	$0.912$  \\
   $\checkmark$ &  &  $\checkmark$ & $1.641$ &	$0.317$ &	$1.048$ & $1.912$ &	$0.251$ &	$0.874$ \\
   \rowcolor{mygray}
    $\checkmark$ & $\checkmark$ & $\checkmark$ & $\bm{1.555}$ &	$\bm{0.322}$ &	$\bm{1.170}$ &	$\bm{1.832}$ &	$\bm{0.261}$ &	$\bm{1.017}$  \\
    \hline
    \Xhline{2.\arrayrulewidth}
    \end{tabular}
\end{table}

\begin{figure*}[t]
    \centering
    \begin{overpic}[width=0.96\linewidth]{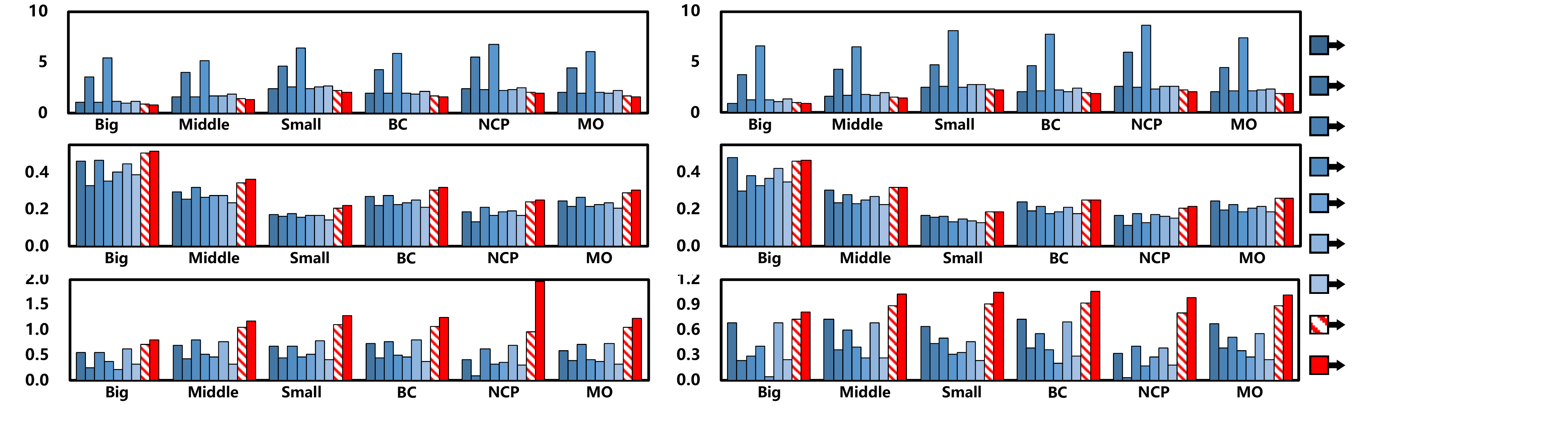}
    
    \put(21.35,-0.2){{\color{black} \small\textbf{Seen}}}
    \put(61.8,-0.2){{\color{black} \small\textbf{Unseen}}}
    
    \put(86,23.9){\tiny\textbf{\citep{kummerer2016deepgaze}}}
    \put(86,21.4){\tiny\textbf{\citep{huang2018predicting}}}
    \put(86,18.7){\tiny\textbf{\citep{mai2020erasing}}}
    \put(86,16.3){\tiny\textbf{\citep{pan2021unveiling}}}
    \put(86,13.73){\tiny\textbf{\citep{gao2021ts}}}
    \put(86,11.32){\tiny\textbf{\citep{BAS}}}
    \put(86,8.6){\tiny\textbf{\citep{nagarajan2019grounded}}}
    \put(86,6.1){\tiny\textbf{\citep{luo2022learning}}}
    \put(86,3.5){\tiny\textbf{Ours}}
    
    \put(-0.1,20.8){{\color{black} \rotatebox{90}{\small\textbf{KLD}}}}
    \put(-0.2,12.7){{\color{black} \rotatebox{90}{\small\textbf{SIM}}}}
    \put(-0.2,4.2){{\color{black} \rotatebox{90}{\small\textbf{NSS}}}}
    \put(-4,22.1){(\bm{$\downarrow$})}
    \put(-4,13.5){(\bm{$\uparrow$})}
    \put(-4,5.){(\bm{$\uparrow$})}
    
    \end{overpic}
    
    \caption{\textbf{The results of the models test with different attributes.} We divide the test set into six subsets according to the attributes described in Sect. \ref{statistic} and test the performance of different models in each of these subsets. The left side shows the results of the experiment in the ``Seen'' setting and the right side shows the results in the ``Unseen'' setting. }
    \label{figure:Attribute}
\end{figure*}
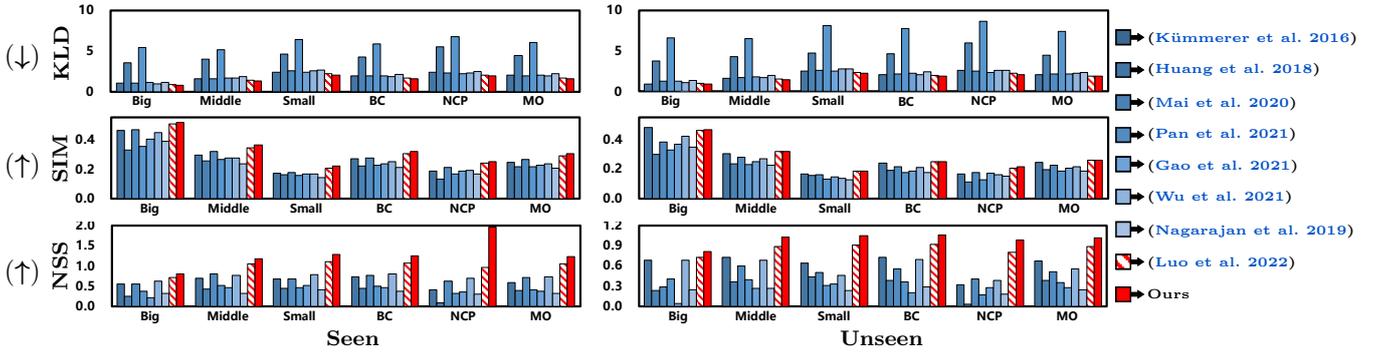

\noindent\textbf{Different Attributes. }Fig. \ref{figure:Attribute} shows the results for all models on different attributes (``Big'', ``Middle'',`` Small'', ``BC'', ``NCP'', and ``MO''). Our approach outperforms other models in almost all settings. In the challenging ``Big'' subset, our method still produces relatively strong results in terms of the NSS metric. Although the results are decreasing in the KLD and SIM metrics, they outperform the other methods. In the three subsets of ``BC'', ``MO'', and ``NCP'', our model achieves superior results to the other models, indicating that our method is more robust and can accurately locate the affordance regions of the object.

\subsection{Ablation Study}
\label{Ablation Study}
\begin{figure*}[t]
	\centering
		\begin{overpic}[width=0.99\linewidth]{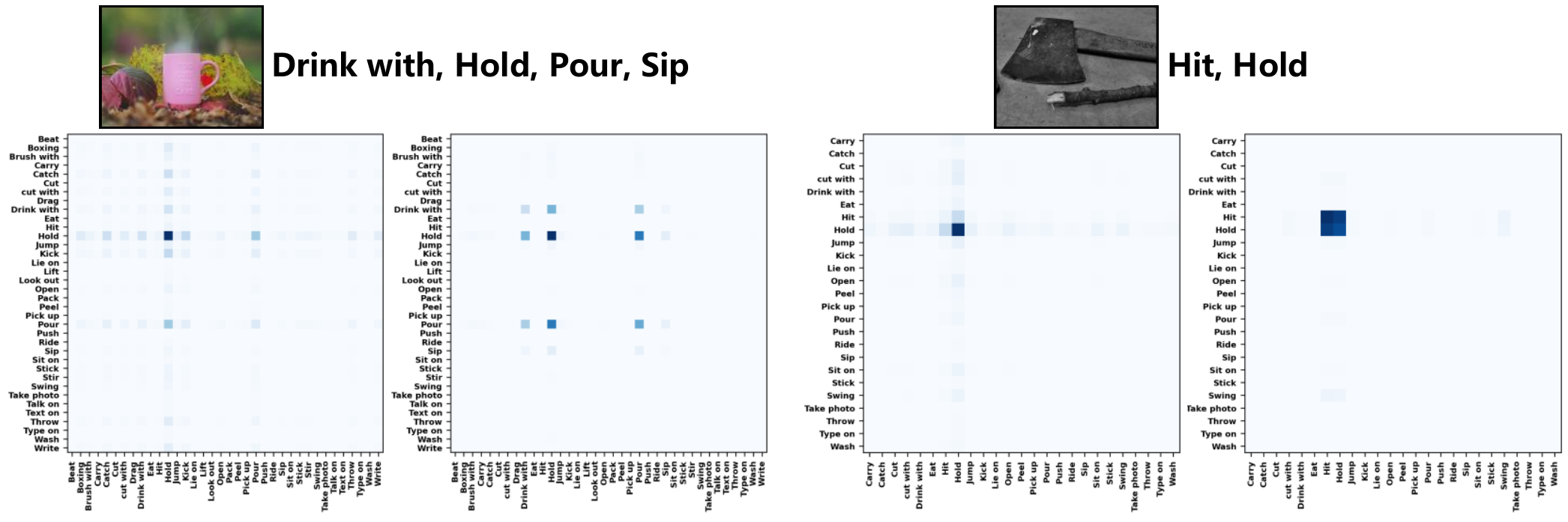}
		\put(10.5,-1.){{\color{black} \small{\textbf{$w$/$o$ ACP}}}}
		\put(36,-1.){{\color{black} \small{\textbf{Ours}}}}
		\put(62.5,-1.){{\color{black} \small{\textbf{$w$/$o$ ACP}}}}
		\put(87.5,-1.){{\color{black} \small{\textbf{Ours}}}}
	\end{overpic}
	\vspace{1pt}
	\caption{\textbf{Visualization of the co-relation matrix.} We visualize the co-relation matrix with or without the ACP strategy.}
	\label{cm}
\end{figure*}

\begin{figure}[t]
	\centering
		\begin{overpic}[width=0.92\linewidth]{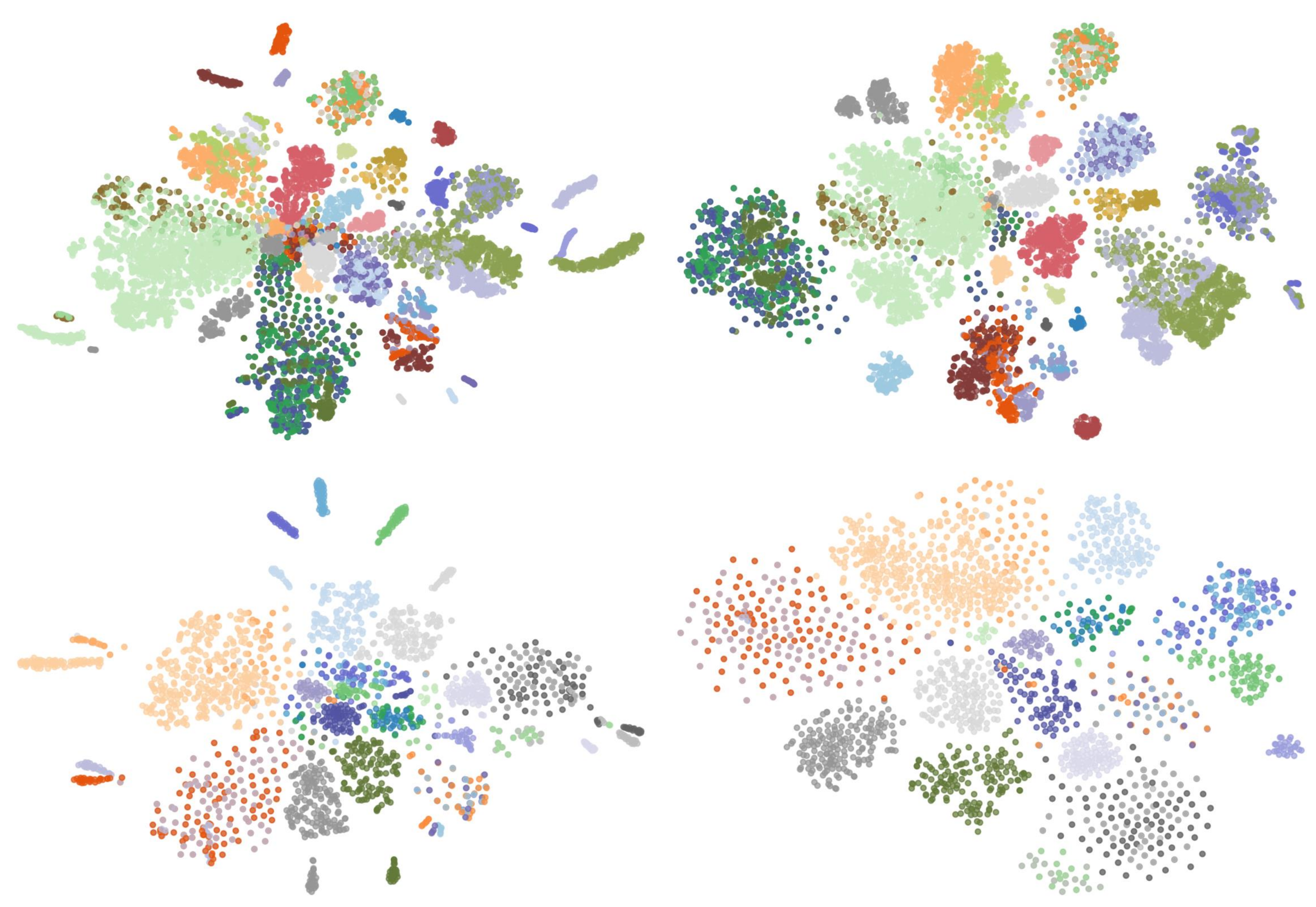}
		\put(17.5,-2.5){{\color{black} \small{\textbf{Baseline}}}}
		\put(70.5,-2.5){{\color{black} \small{\textbf{Ours}}}}
		
		\put(-3.,45){\rotatebox{90}{{\color{black} \small{\textbf{Seen}}}}}
		\put(-3.,9){\rotatebox{90}{{\color{black} \small{\textbf{Unseen}}}}}
	\end{overpic}
	\vspace{3pt}
	\caption{\textbf{T-SNE visualization results.} The T-SNE results for the baseline without any modules and our model. The first row is the ``Seen'' setting and the second row is the ``Unseen'' setting. Due to the small number of images in the ``Unseen'' setting, the results in the second row are sparse.}
	\label{tsne}
\end{figure}
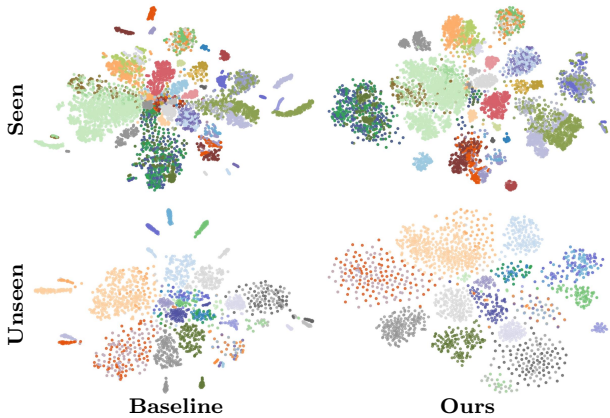

\begin{figure}[t]
	\centering
		\begin{overpic}[width=0.98\linewidth]{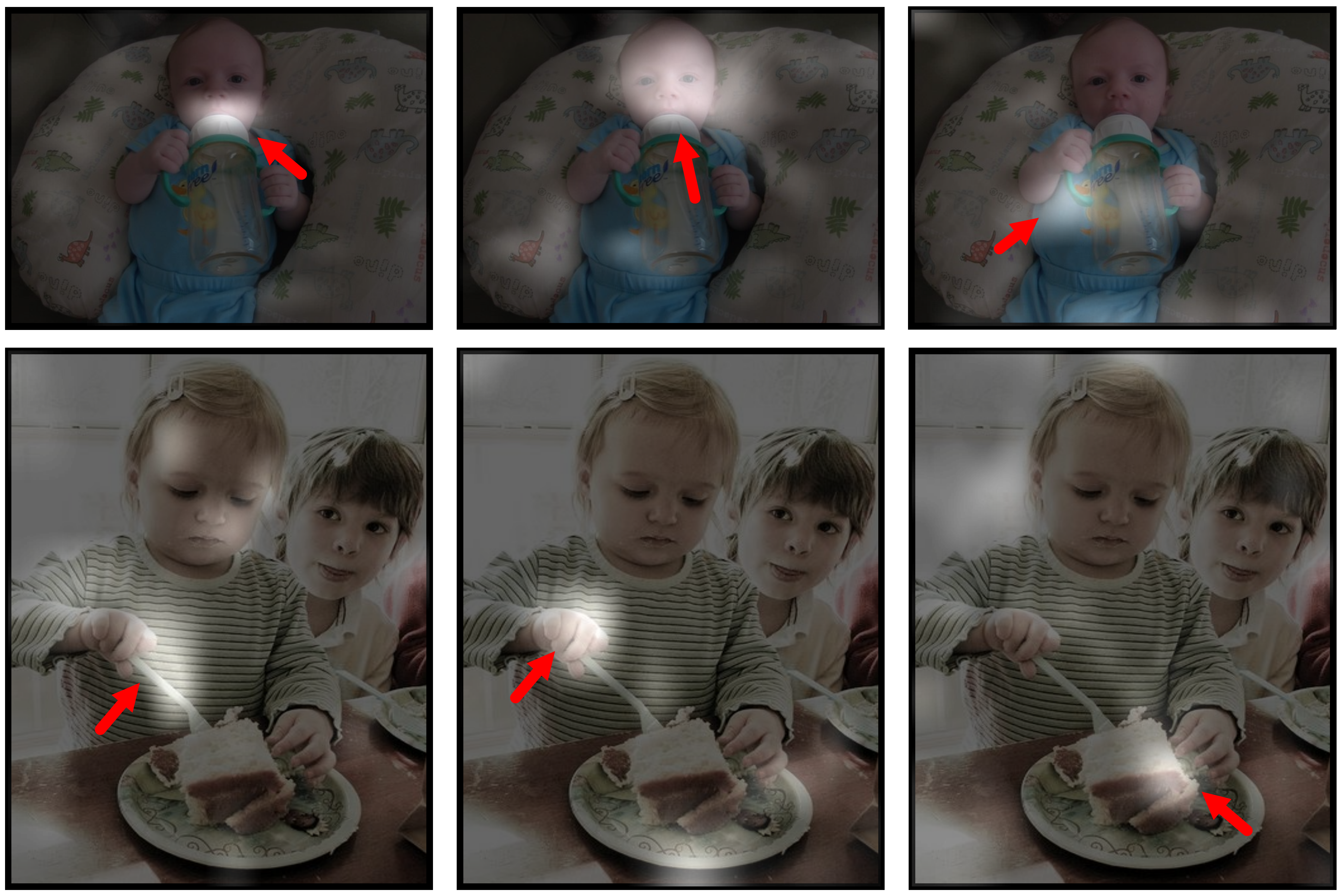}
	\end{overpic}
	\vspace{-3pt}	\caption{\textbf{Visualization of the coefficient matrix $H$ for the last iteration of the AIM module.} We choose maps activated by different bases.}
	\label{result_exo}
\end{figure}

\begin{figure}[t]
	\centering
		\begin{overpic}[width=0.98\linewidth]{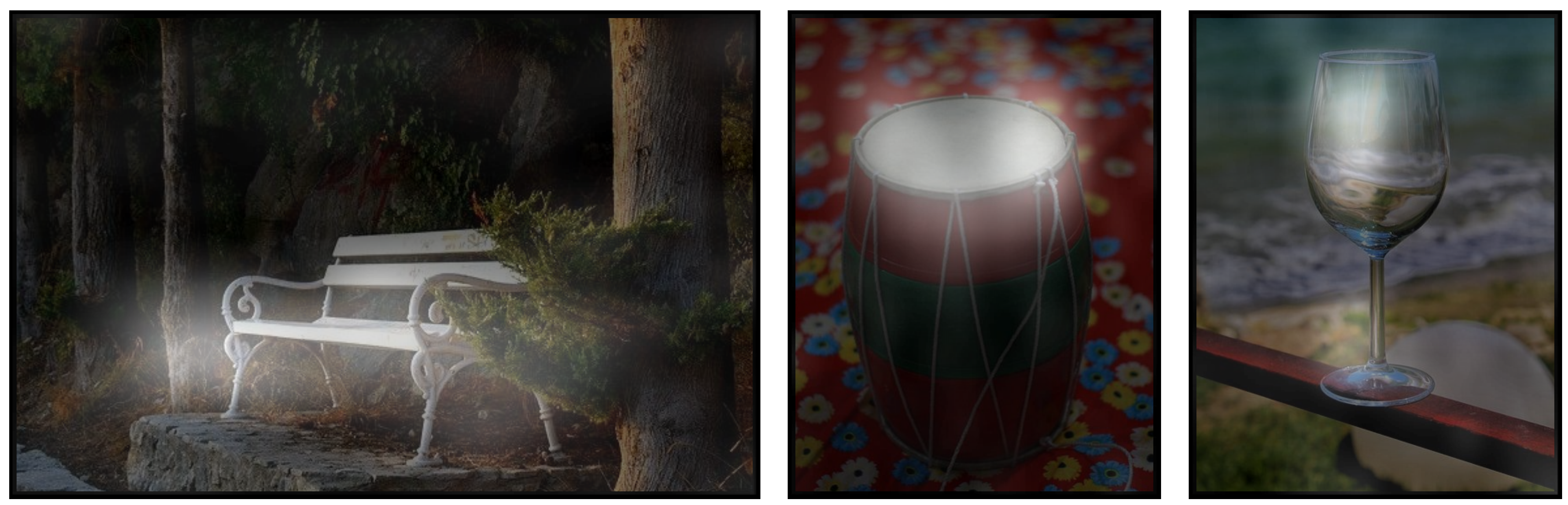}
       \vspace{-1pt}
	\end{overpic}
	
	\caption{\textbf{Visualization of the CFT module outputs.}}
	\label{result_ego}
\end{figure}

To investigate the impact of the AIM module, the CFT module, and the ACP strategy, we evaluate all combinations, as shown in Table \ref{Table:ablation study1}. Note that since the matching of affordance regions in the CFT module requires an optimized dictionary base matrix in the AIM module, the CFT module's presence must depend on the AIM module. It indicates that the AIM module extracts invariant affordance features from the diverse interactions of exocentric views, which enables the model to extract affordance-related features quickly. Meanwhile, the ACP strategy can enhance the perception of the affordance region by aligning the co-relation of the two branches. We make the T-SNE feature visualization of our model and baseline, as shown in Fig. \ref{tsne}. It shows that our method has more explicit discriminative boundaries and can accurately distinguish different affordance features. To analyze the features mined by the AIM module from multiple exocentric images, we visualized the correlation coefficient matrix $H$ of the AIM module (as shown in Fig. \ref{result_exo}). The maps represent the attention maps generated by different dictionary bases in $W$, indicating that different dictionary bases focus on different regions during human-object interaction. When the human drinks, the AIM module can focus on the region where the mouth is in contact with the bottle. The activated face can provide valuable contextual clues for reasoning about the action of ``Drink''. At the same time, there is a co-relation between ``Drink'' and ``Hold'', so the AIM module also activates the region of hand interaction. It indicates that different AIM module bases focus on interaction-related features and jointly form a better representation. To verify the ability of the CFT module to transfer affordance-related features to egocentric features, we visualize the output features of the CFT module. We take the mean value of the channels for egocentric features and obtain the results shown in Fig. \ref{result_ego}. The CFT module can locate the interaction regions corresponding to affordance, thus providing more reliable features for cross-view knowledge transfer. To evaluate the ACP strategy for object availability co-relation mining, we visualize the co-relation matrix of the output of the egocentric branch in the testing phase separately (shown in Fig. \ref{cm}). The left figure shows that our model could better capture the co-relation between the affordance classes ``Drink with'' and ``Hold'' of the cup, and suppress irrelevant category predictions. The right figure shows that our approach can explore the possible existence of multiple affordance classes (``Hold'' and ``Hit'') in the same interaction region.

\begin{figure*}[t]
    \centering
    \small
    \begin{overpic}[width=0.98\linewidth]{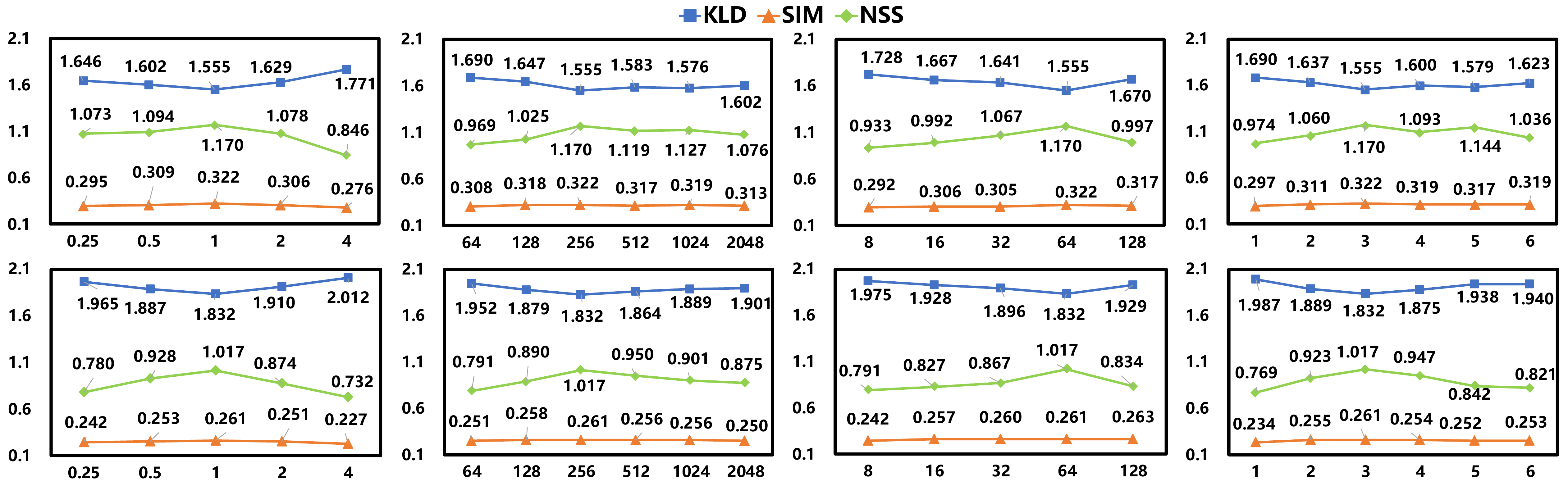}
    \put(12.5,-1.3){{\color{black} \textbf{(a)}}}
    \put(37.3,-1.3){{\color{black} \textbf{(b)}}}
    \put(62.6,-1.3){{\color{black} \textbf{(c)}}}
    \put(87.5,-1.3){{\color{black} \textbf{(d)}}}
    \end{overpic}
    \vspace{5pt}
    \caption{\textbf{Different Hyper-parameters.} We explore \textbf{(a)} the effects of different $T$ values in the ACP strategy, \textbf{(b)} different number of channels $c$ in the input features in the AIM module, \textbf{(c)} different rank $r$ of the dictionary matrix in the AIM module, and \textbf{(d)} different number of exocentric images $N$ on the model performance. The first and second rows show the results of the experiments with the ``Seen'' and ``Unseen'' settings respectively.}
    \label{figure:Different Hyper-parameters}
\end{figure*}

Fig. \ref{figure:Different Hyper-parameters} (a) shows the influence of $T$ in the ACP strategy on the model. $T$ has a smoothing effect on the category correlation distribution and plays a preservation role for affordance co-relation. The performance is more sensitive to changes in $T$ and has a greater impact at larger values. Fig. \ref{figure:Different Hyper-parameters} (b) and (c) show the effect of the channel dimension $c$ of the features and the rank $r$ of the dictionary matrix $W$ in the AIM module, respectively. The value of $c$ has no significant impact on the results, and the model achieves a slightly better performance at $c=\text{512}$. A larger number of channels may increase the complexity of the optimization and lead to a decrease in model performance. Different ranks represent the number of bases of the interaction subfeatures. The best results are obtained when $r=\text{64}$. A smaller $r$ (for example $r=8$) may lead to poor results due to the number of bases being too small to represent the interactions' sub-features fully. Moreover, a larger $r$ leads to worse results possibly due to the redundancy of information caused by redundant bases. Fig. \ref{figure:Different Hyper-parameters} (d) shows the impact of the number of exocentric images on model performance, which has a relatively positive effect on model performance as $N$ increases from 1 to 3. It indicates that the AIM module can capture affordance-specific cues from multiple images, playing a critical role in affordance region prediction.

\begin{figure}[t]
    \centering
    \begin{overpic}[width=0.99\linewidth]{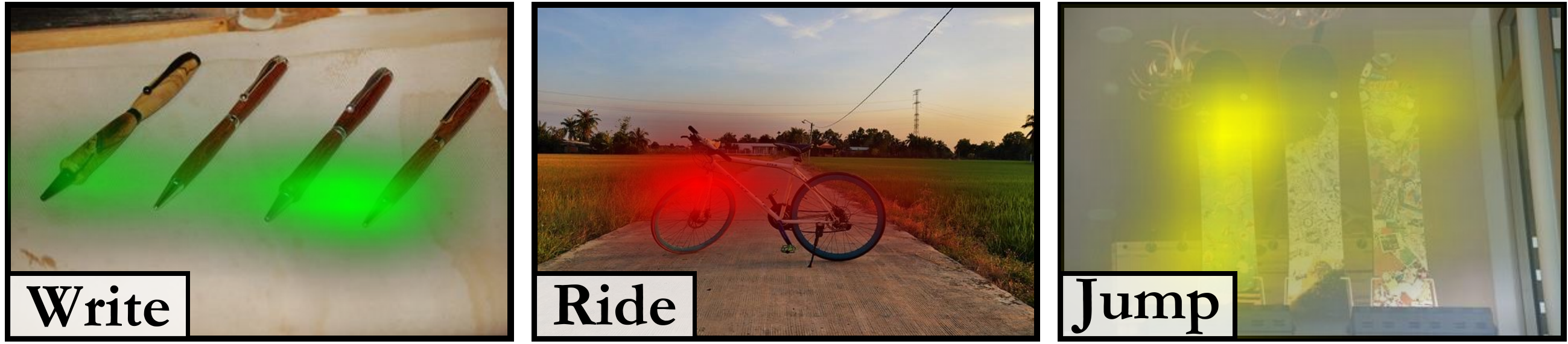}
    \end{overpic}
    \caption{\textbf{Failure cases.} Some examples of the model's failure in slender structures, complex structures and indistinct front and back views.}
    \label{figure:FC}
\end{figure} 

\begin{figure}[t]
    \centering
    \begin{overpic}[width=0.99\linewidth]{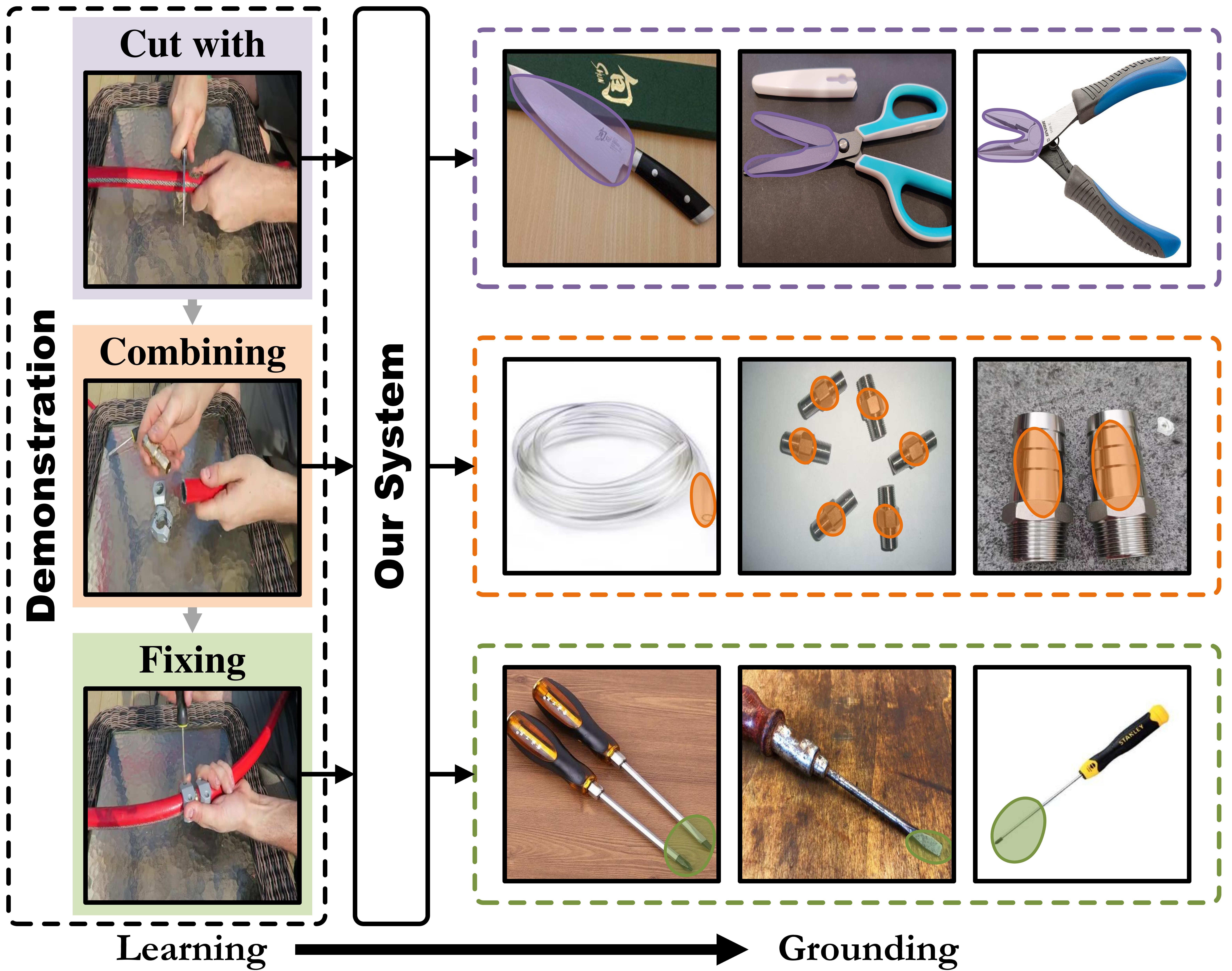}
    \end{overpic}
   
    \caption{\textbf{Potential Applications: Learning from demonstrations.} Our system can quickly extract the interaction region from the human demonstration image and locate it in the egocentric image.}
    \label{figure:app1}
\end{figure} 

\begin{figure}[t]
    \centering
    \begin{overpic}[width=0.99\linewidth]{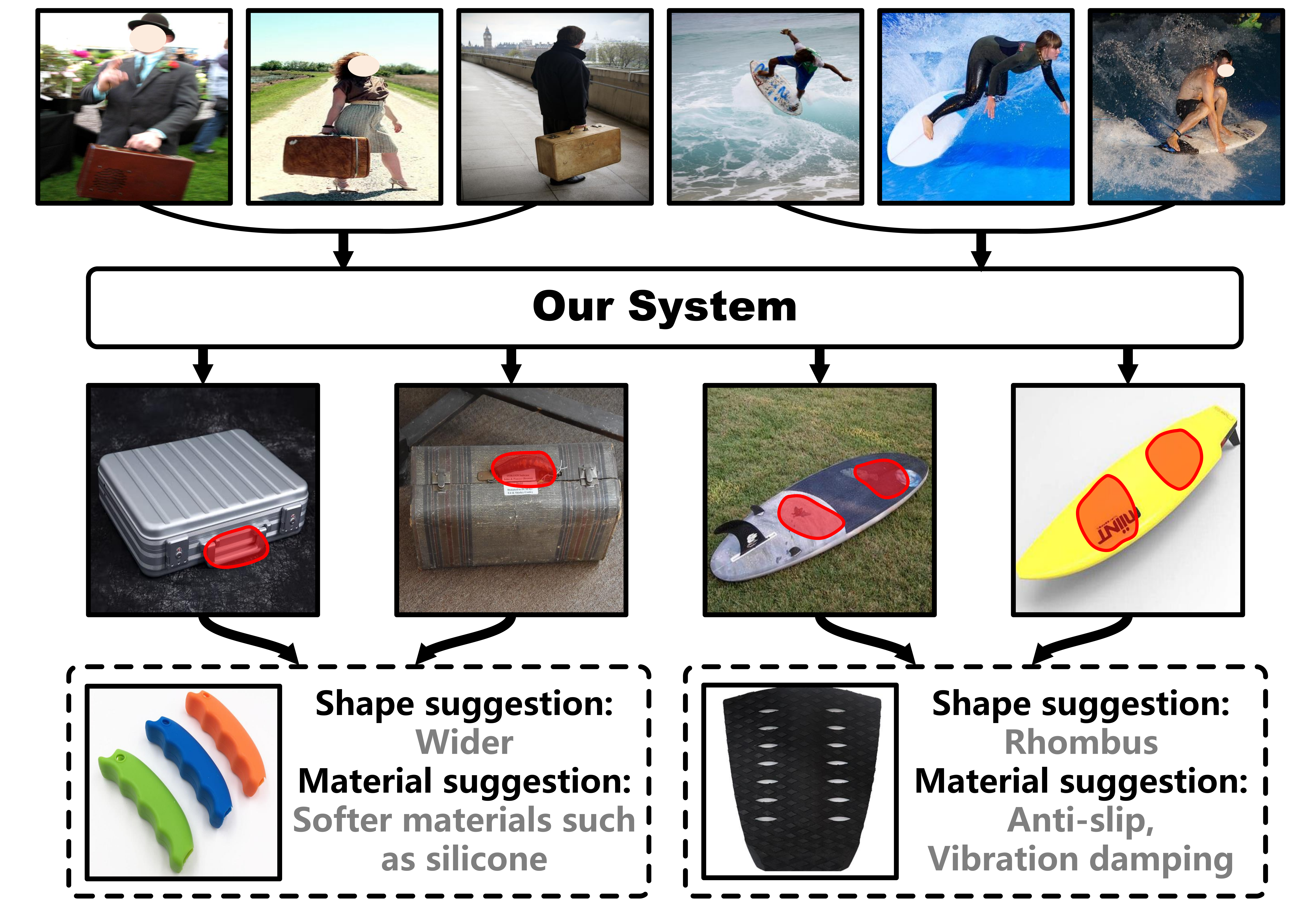}
    \end{overpic}
    \caption{\textbf{Potential Applications: Industrial Manufacturing.} Our system provides the ability to help design the product's shape and material according to the interaction habits during the manufacturing process.}
    \label{figure:app2}
\end{figure}

\section{Conclusion and Discussion}
\label{sec:Conclusion and Discussion}
In this paper, we make the first attempt to address a challenging task named affordance grounding from the exocentric view. Specifically, we propose a novel cross-view affordance knowledge transfer framework that can extract invariant affordance from diverse exocentric interactions and transfer it to an egocentric view. We also establish a large affordance grounding dataset named AGD20K, which contains 20K well-annotated images, serving as a pioneer testbed for the task. Besides, we expand the scale of the test set from a wide range of different attributes, making the test set more challenging and more suitable for real-world application scenarios. Our model outperforms eight representative models from four related areas and can serve as a strong baseline for future research. 

\noindent\textbf{Future Directions. }Future research could focus on more precisely localizing human body parts (such as the hands, feet, mouth, \etc) to interact with objects and recognizing each local region that the human body interacts with based on an exocentric view. Moreover, object affordance area localization could be studied in multimodal scenarios which contain language or audio data.  Additionally, exploring the development of various prompts employing large models \citep{kirillov2023segment,shen2023hugginggpt,ramesh2022hierarchical,zhou2022learning} to enhance the identification of an object's affordance with greater efficiency is worth examining.

\noindent\textbf{Weakness. }Fig. \ref{figure:FC} shows some failure cases. Our model may activate irrelevant background when the structure is lengthy, thin, and complex or when the background and foreground cannot clearly distinguish. Future work could focus on enhancing the associated affordance class regions during training, ignoring the irrelevant background regions \citep{BAS}, and adjusting the obtained results according to affordance properties to produce more accurate predictions \citep{pan2021unveiling}. 

\noindent\textbf{Potential Applications. }\textbf{1) Learning from demonstrations:} As shown in Fig. \ref{figure:app1}, our system empowers the agent to efficiently acquire affordance-related knowledge from the human-object interaction in the exocentric view and to locate it in the egocentric image, which enables the agent to operate in the first-person view \citep{fang2018demo2vec,nagarajan2019grounded}.
\textbf{2) Industrial Manufacturing:} As shown in Fig. \ref{figure:app2}, our system can explore the object's interaction regions from multiple exocentric views of the human-object interaction and design more suitable shapes and materials to improve product quality during the manufacturing process \citep{lau2016tactile}.

\bibliographystyle{spbasic}
\bibliography{ijcv}

\end{document}